# Unveiling the frontiers of deep learning: Innovations shaping diverse domains

Shams Forruque Ahmed[1,2,*], Md. Sakib Bin Alam[2], Maliha Kabir[4], Shaila Afrin[4], Sabiha Jannat Rafa[4], Aanushka Mehjabin[5], Amir H. Gandomi[6,7,8,*]

[1] School of Mathematical Sciences, Sunway University, Bandar Sunway, Petaling Jaya 47500, Selangor Darul Ehsan, Malaysia
[2] Department of Mathematics & Physics, North South University, Dhaka 1229, Bangladesh
[3] Department of Information Technology, University of Information Technology and Sciences, Dhaka 1212, Bangladesh
[4] Science and Math Program, Asian University for Women, Chattogram 4000, Bangladesh
[5] School of Biological Sciences, Georgia Institute of Technology, Atlanta, GA 30332, USA
[6] Faculty of Engineering & Information Technology, University of Technology Sydney, NSW, 2007, Australia
[7] University Research and Innovation Center (EKIK), Óbuda University, 1034 Budapest, Hungary
[8] Department of Computer Science, Khazar University, Mahsati 41, Baku, Azerbaijan

[*]Corresponding authors: shams.forruque@northsouth.edu, shams.f.ahmed@gmail.com (Shams Forruque Ahmed); gandomi@uts.edu.au (Amir H. Gandomi)

## Abstract

Deep learning (DL) allows computer models to learn, visualize, optimize, refine, and predict data. To understand its present state, examining the most recent advancements and applications of deep learning across various domains is essential. However, prior reviews focused on DL applications in only one or two domains. The current review thoroughly investigates the use of DL in four different broad fields due to the plenty of relevant research literature in these domains. This wide range of coverage provides a comprehensive and interconnected understanding of DL's influence and opportunities, which is lacking in other reviews. The study also discusses DL frameworks and addresses the benefits and challenges of utilizing DL in each field, which is only occasionally available in other reviews. DL frameworks like TensorFlow and PyTorch make it easy to develop innovative DL applications across diverse domains by providing model development and deployment platforms. This helps bridge theoretical progress and practical implementation. Deep learning solves complex problems and advances technology in many fields, demonstrating its revolutionary potential and adaptability. CNN-LSTM models with attention mechanisms can forecast traffic with 99% accuracy. Fungal-diseased mango leaves can be classified with 97.13% accuracy by the multi-layer CNN model. However, deep learning requires rigorous data collection to analyze and process large amounts of data because it is independent of training data. Thus, large-scale medical, research, healthcare, and environmental data compilation are challenging, reducing deep learning effectiveness. Future research should address data volume, privacy, domain complexity, and data quality issues in DL datasets.

**List of abbreviations**

| | | | |
|---|---|---|---|
| AI | Artificial intelligence | HDFS | Hadoop Distributed File System |
| AMP | Antimicrobial peptide | HPE | Human pose estimation |
| ANN | Artificial neural network | LDA | Linear discriminant analysis |
| API | Application programming interface | MCA | Multiple correspondence analysis |
| ATS | Automatic Text Summarization | MDNN | Multimodal deep neural network |
| Bi-LSTM | Bidirectional-LSTM | MFCC | Mel Frequency Cepstral Coefficients |
| CAN | Collaborative adversarial network | ML | Machine learning |
| CDN | Convolutional decoder networks | NLP | Natural language processing |
| CEN | Convolutional Experts Network | NMT | Neural machine translation |
| CFD | Computational fluid dynamics | OTE | Opinion target expression |
| CNN | Convolutional neural network | PCA | Principal component analysis |
| CNTK | Computational Network Tool Kit | PRM | Pyramid Residual Module |
| CPU | Central processing unit | QA | Question answering |
| DAM | Domain-attention model | RL | Reinforcement learning |
| DDI | Drug-drug interaction | RNN | Recurrent neural network |
| DL | Deep learning | ROM | Reduced-order modeling |
| DMLC | Distributed Machine Learning Common | SER | Speech emotion recognition |
| ECs | Emerging contaminants | SHBG | Sex-hormone binding globulin |
| EM | Electron microscopy | SHM | Structural health monitoring |
| EPI | Enhancer-promoter interactions | SNN | Spiking neural network |
| ER | Estrogen receptor | SNR | Signal-to-noise ratio |
| FBE | Filter bank energies | SVM | Support vector machine |
| GMM | Gaussian mixture model | TEO | Teager Energy Operator |
| GPU | Graphics processing unit | TFX | TensorFlow Extended |
| HAR | Human activity recognition | TPU | Tensor processing unit |
| HDF | Hierarchical Data Format | TSS | Text summarization embedding space |
| LMT | Logistic Model Tree | WS-MDL | Weakly-supervised multimodal deep |
| LR | Linear regression | | learning |
| LSTM | Long short-term memory | | |

## 1. Introduction

Deep learning (DL) is a method for constructing computational models composed of numerous processing layers in order to investigate and learn the demonstrations of multiple abstraction data [1]. Utilizing these pathways, improvements have been made in various sectors, including audio-visual technology, better recognition and detection, genomics, proteomics, biomedicine, drug discovery, environment, and security. Deep learning unravels and identifies the underlying structures within massive datasets utilizing several algorithms, such as back-propagation, to learn and apply changes to given conditions as a machine would. Deep learning exhibits advantages over earlier machine learning (ML) and artificial intelligence (AI) algorithms that lack the ability to analyze natural raw data. The origins of DL can be traced back to the 1940s, when the notion of artificial neural networks was developed, drawing inspiration from the complicated structure and functioning of the human brain [3]. Nevertheless, the proliferation of backpropagation algorithms and the accessibility of more extensive datasets were crucial factors in the 2000s that propelled deep learning to prominence [4]. The advancements during the 2010s propelled deep learning to the forefront of artificial intelligence research and application [5], such as introducing deep learning frameworks like TensorFlow and PyTorch and utilizing deep neural networks for image classification. Multiple statistical trends demonstrate the incredible expansion of deep learning. The deep learning industry has grown rapidly, with a market size of billions of dollars and a substantial compound annual growth rate. The broad use of deep learning in many sectors, such as healthcare, automotive, finance, and technology, fuels this expansion. The need for high-performance hardware, like graphics processing units (GPUs) and tensor processing units (TPUs), to facilitate the training and deployment of complicated neural networks highlights the continued significance of computational power. Statistical trends like these highlight how dynamic and influential deep learning is becoming as its applications spread across more and more domains.

By utilizing methods like representation learning, a system is able to take raw data as input and determine the necessary patterns for analysis. Deep learning takes into account many layers of representation data, each of which affects the other. It is essential to examine the recent advancements and applications of deep learning in various fields. Recent progress in deep learning has driven significant advancements in various domains, from healthcare to finance. For instance, deep learning is improving medical diagnostics by utilizing image recognition, and it is transforming fraud detection and portfolio optimization in the field of finance. This demonstrates the wide-ranging influence and adaptability of this technology. The potential fields of DL applications include audio-visual data processing [6–8], agriculture [9–13], transportation prediction [14–16], natural language processing [17–19], biomedicine [20–22], disaster management [23–25], bioinformatics [26–28], healthcare [29–31], drug design [32–36], financial fraud detection [37, 38], computer vision [39–41], ecology [42–44], fluid dynamics [45–47], and civil engineering [48–50], to understand the present state of deep learning. The studies [6-50]  reviewed applications of deep learning in a particular domain, while others concentrated on specific topics within that discipline. Most of them offered perspectives on DL application in their respective fields by discussing its advantages and disadvantages in light of future developments. Although various forms of analysis were made in many studies, the specific technical aspects of DL applications were not addressed. For instance, Litjens et al. [35] adopted a broad approach to medical image segmentation and reviewed many different subfields, which did not help to limit the topic. Hesamian et al. [36] shed light on the machine learning and artificial intelligence techniques utilized in recent biomedical image research, focusing on their structure and methodology and evaluating their benefits and limitations. They did not discuss the difficulties of black boxes and the inaccessibility of complex neural networks to human cognition. Getting a complete picture of deep learning's capabilities and limitations is challenging if reviews focus on one study area, as DL models may have issues generalizing across different domains. Additionally, this limited scope fails to account for the more extensive ethical, societal, and interdisciplinary factors vital for a comprehensive assessment of the impacts of deep learning. Therefore, the present investigation is undertaken to address these gaps by taking into account all of the aforementioned potential disciplines.

Several studies [51–54] evaluated features of deep learning-based audio-visual data processing, such as text analysis, activity recognition, and face recognition, without exploring the difficulties of black boxes or data privacy issues. Particularly, in sampling text and picture data, the processing of personally identifiable information is subject to numerous legal constraints that are not adequately clarified. Sorting and analyzing biomedical data are crucial challenges in the health industry since it is complicated, varied, and dispersed. Health records, picture records, genomes, proteomics, transcriptomics, sensory data, and texts are just a few examples of the many types of data produced by the biomedical sector. To accurately forecast, represent, analyze, and sort this data, deep learning-based data mining approaches have proven to be both effective and fast. The usage of spiking neural networks (SNNs), which are constructed to emulate the information processing methods in biological systems, was reviewed by Pfeiffer et al. [55]. However, the study failed to address the issues posed by these systems, such as the fact that massive volumes of data input cause delays and fail to incorporate uncertainties. Many other reviews concentrating on deep learning techniques for health data overlook the challenges posed by poor data quality and excessive data volume [56]. Privacy problems of health data and genomic data are commonly disregarded in data mining approaches in biomedicine and bioinformatics [28, 57]. Yuan et al. [58] reviewed conventional neural network and deep learning approaches, which pertain to the development of environmental remote sensing procedures and their applications in environmental monitoring. Nevertheless, the authors did not explain how to deal with issues like incomplete or inaccurate data, complex domains, inaccurate models, or the difficulty of using a multidisciplinary approach to merge disciplines.

A search of the scholarly literature was performed to identify available studies on deep learning applications across numerous sectors, deep learning frameworks, and their associated challenges and benefits, utilizing an integrative approach, which comprises a broad collection, careful screening, and intensive assessment of pertinent and high-caliber articles. Searches were carried out utilizing databases sourced from reputable organizations such as Web of Science, Scopus, and Google Scholar, and well-reputed publishers like IEEE, De Gruyter, Elsevier, Nature, Taylor & Francis, Springer, ACM, Wiley, World Scientific, Inderscience, Oxford University Press, Emerald, IOP, Cambridge University Press, and Sage. The most pertinent publications for this study were identified through a search utilizing the following keywords: Deep learning, deep learning framework, deep learning application, use of deep learning, applications of deep learning, challenges/limitations/disadvantages of deep learning, benefits/advantages of deep learning, deep learning in a particular discipline, and similar terms. Then, by collecting and filtering the bibliographies and references of the aforementioned publications, additional relevant papers were identified. Through the search, it has been found that there have been many recent reviews [6, 10, 14, 18, 21, 23, 59-69] of deep learning applications. However, as seen in Table 1, the reviews focused primarily on applications of deep learning in just one or two fields. In addition, there is currently no available review that investigates all possible applications of deep learning. For the advancement in deep learning research, it is crucial to understand and compare deep learning's performance across different fields. In order to assess the performance, application, progress, and challenges associated with deep learning, this study conducts a comprehensive review of its potential applications. This review took into account four different fields because there is plenty of relevant research literature in these areas where deep learning is widely applied. This report will serve as a valuable resource for academics and industrialists, offering comprehensive insights into the diverse practical applications of deep learning in various industries. Additionally, it will provide a deeper understanding of the potential challenges encountered during the deep learning implementation process.

Table 1. Comparison between the present and the most recent review studies on deep learning applications

| Review study | Main task (s) | Deep learning frameworks | No. of DL application fields | Challenges and benefits of DL |
|---|---|---|---|---|
| This study | Investigation of DL applications across all major fields of study, along with their associated benefits and challenges | √ | √ (4) | √ |
| Zhu et al. [6] | Examine recent developments in audio-visual learning. | × | √ (1) | √ |
| Wang et al. [10] | Explore DL utilization for analyzing hyperspectral images in agriculture | ×. | √ (1) | × |
| Wang et al. [14] | Investigate the DL techniques to improve the cognitive capabilities of transportation systems. | × | √ (1) | × |
| Otter et al. [18] | Advances in natural language processing using deep learning | × | √ (1) | × |
| Haque and Neubert [21] | Implementation of deep learning techniques for biomedical image segmentation | × | √ (1) | × |
| Sun et al. [23] | Provides an overview of DL applications in disaster management | × | √ (1) | × |
| Shadmani et al. [59] | Predicting wave energy resources and optimizing the design of wave energy converter with the help of ML and DL | × | √ (2) | × |
| Kipli et al. [60] | Elucidate the utilization of deep learning techniques and their practical application in the domain of oil palm tree counting. | × | √ (1) | √ |
| Huang et al. [61] | Applications of DL in dentistry, with an emphasis on dental imaging | × | √ (1) | × |
| Erfanian et al. [62] | DL applications in genomics and transcriptomics | × | √ (2) | × |
| Adam et al. [63] | Use of DL to detect breast cancer based on the basis of magnetic resonance imaging | × | √ (1) | × |
| Lata and Cenkeramaddi [64] | Investigate the performance of DL methods in medical image cryptography to protect sensitive patient information. | × | √ (1) | × |
| Hederman and Ackerman [65] | Using DL to enhance vaccine development | × | √ (1) | √ |
| Yang et al. [66] | Outline the pathophysiological findings of pathogenic variants in neurological disorders' coding and noncoding transcripts using DL. | × | √ (1) | √ |
| Ruszczak et al. [67] | Applying DL to power grid diagnostics | × | √ (1) | × |

| Wani et al. [68] | Detection of brain tumors through the utilization of deep neural networks | × | √ (1) | √ |
| Alqahtani et al.[69] | The accuracy of dyslexia prediction using deep learning models | × | √ (1) | √ |

×: Not availbale; √ : Available

The paper is structured as follows: Different types of deep learning frameworks are introduced and described in Section 2. Section 3 analyzes and explores the applications of deep learning in different sectors. The challenges of different DL application fields are summarized in Section 4, along with their benefits. Section 5 outlines the potential future research in order to minimize the identified research gaps in deep learning. A concluding review of the study is presented in Section 6.

## 2. Deep learning frameworks

Deep learning techniques can be referred to as a representation learning method, which has many layers of representation. These representation components can be created by building non-linear elements that can successively change the degree of representation into a subsequent abstract level. Deep learning can perform very complicated functions by combining these modifications. Many deep learning frameworks are used to develop solutions for complex problems, such as Caffe/Caffe2 [70–72], MXNet [73–76], Computational Network Tool Kit (CNTK) [77, 78], Torch/PyTorch [74, 75, 79, 80], Theano [81, 82], Chainer [83, 84], PaddlePaddle [85–87], Keras [75, 88, 89], and TensorFlow [74, 89–91], which are discussed in this section. **Fig. 1** shows several frameworks of deep learning.

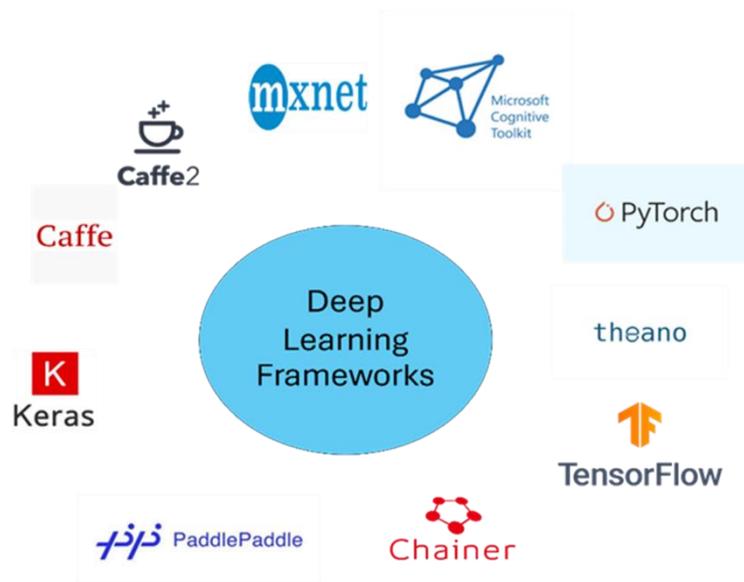

**Fig. 1.** Deep learning frameworks

## 2.1 Caffe/Caffe2

Caffe stands for convolutional architecture for rapid feature embedding. It is a free and open-source DL framework that allows users to explore complex structures. This library was created in C++ by the BVLC center and is often used in Matlab and Python [92]. Caffe utilizes declarative configuration for defining neural network designs, and it enables several layer types, such as fully connected, convolutional, and pooling layers. Both Caffe and Caffe2 are on the basis of feedforward neural networks, which use a non-linear activation function after a linear transformation in each layer [92]. The forward pass can be mathematically expressed as:

$$h^{(l)} = f(W^{(l)} h^{(l-1)} + b^{(l)})$$
(1)

where $h^{(l)}$, $W^{(l)}$, and $b^{(l)}$ represent the activation vectors at the layer $l$; $f \cdot$ denotes the activation function such as sigmoid, ReLU, and tanh.

Caffe2 is an advanced version of Caffe created by Yangqing Jia, the same person who created Caffe. After Yangqing Jia began operating for Facebook, they collaborated with Facebook and NVIDIA to develop the Caffe2 framework based on Caffe [93]. Caffe can handle over forty million pictures daily with only a Titan GPU or K40 [94]. In Caffe, data are received by data layers. Because of the standard picture formats of these data layers (.gif, .tiff, .jpg, .jpeg, .png, .pdf), Hierarchical Data Format (HDF5) and efficient databases are acceptable. If Caffe can be incorporated with cuDNN, it will enhance the productivity rate by 36% without using too much memory [95]. Although more recent frameworks have surfaced featuring distinct design philosophies, Caffe continues to be a valuable instrument for DL model development tasks that require simplicity and speed. Caffe2 addresses several of Caffe's shortcomings, including vast dispersed training, deployment in mobile systems, and additional equipment support for quantized processing. NVIDIA is an excellent support system for Caffe2, which includes Python and C++ APIs [96]. This enables it to prototype and refine projects quickly.

Caffee and Caffe2 also have some limitations. Caffe's lack of thorough documentation makes it difficult for new users to get going or troubleshoot problems. Because of its more strict design, implementing custom layers or experimental architectures may be difficult. Its poor performance with big neural networks and recurrent neural networks (RNNs) causes complicated models to be handled laboriously and inefficiently. Caffe's development has slowed considerably, restricting its potential and rendering it less appropriate for contemporary applications that demand constant innovation. On the other hand, new users may find compatibility challenging in Caffe2 due to changes from Caffe. Its versatility for some applications is limited by its lack of support for dynamic graph calculations.

## 2.2 MXNet

MXNet is a multilingual deep learning framework made to achieve speed and flexibility [74]. Pedro Domingos and a team of scientists developed MXNet, which is also an addition of DMLC (Distributed (Deep) Machine Learning Common). This framework is highly scalable with a small memory footprint. The MXNet is constructed using tensor operations and has the capability to handle dynamic computational graphs. The primary function is the convolution, which can be represented as:

$$Y = W * X + b$$
(2)

In Eq. (2), $X$ and $Y$ denote the input and output tensors, respectively; $W$ signifies filter tensor, $*$ indicates convolution operation; $b$ denotes bias tensor.

MXNet can operate on a vast range of platforms, including GPU systems and mobile devices, and can perform numerical calculations with a short code of Python and R for dispersed networks and GPUs. It supports imperative and symbolic programming using NDArray API and Symbol API [97]. The MXNet platform offers extensive tools and libraries for constructing and implementing complex neural network models in diverse fields such as natural language processing, computer vision, and reinforcement learning. The dynamic computation graph of MXNet facilitates convenient experimentation and fast prototyping, rendering it highly favored by both researchers and practitioners. MXNet, with the active participation of a lively community, continues to develop and adapt, ensuring its competitiveness in the ever-changing field of deep learning frameworks.

Despite these strengths, MXNet's limitations include compatibility issues with certain APIs, which might make integrating with other tools and libraries difficult. This can be a major disadvantage in large data pipelines or multi-framework ecosystems where smooth system-to-system communication is essential. Differences in design principles, version mismatches, or inadequate support for more recent features in third-party libraries can all lead to compatibility problems. MXNet provides dynamic graph computing and is compatible with many computer languages; nevertheless, new users may find the learning curve challenging. Furthermore, MXNet might not be as intuitive as other frameworks, making it less suitable for novices.

### 2.3 Computational network tool kit

Computational network tool kit (CNTK) is a deep learning framework that features Python API and is built on top of C++ code developed by Microsoft [98]. This framework represents neural networks as a sequential computational process. CNTK is renowned for its efficacy in deep neural network training and is specifically engineered to manage complex, extensive tasks. CNTK makes use of computation networks, in which nodes stand for operations and edges for data dependencies [98]. Matrix multiplication following an activation function is a common operation in CNTK:

$$y = \sigma(Wx + b) \tag{3}$$

where $x$ and y represent the input and output vectors, respectively; $W$ denotes the weight matrix; $b$ signifies the bias vector; $\sigma(\cdot)$ defines the activation function.

CNTK's distinctive feature is its seamless integration with the deep learning tools and services offered by Microsoft, which positions it as a favored option among users operating within the Microsoft ecosystem. CNTK can readily combine some of the prominent architectures like Feed-forward deep neural networks (DNNs), convolutional neural networks (CNNs), and recurrent neural networks (RNNs). C# and BrainScript are supported by CNTK, which both offer increased and minimal APIs for simplicity of usage and versatility. By providing a performance-optimized backend and supporting acceleration on both CPU and GPU, CNTK is well-suited for various applications, including natural language processing (NLP) and speech and image recognition. For instance, CNTK was evaluated in a study using a system of several GPUs in a thoroughly linked four-layer neural network [99] and was shown to outperform Caffe, TensorFlow, Theano, and Torch [100].

Because of its complexity, CNTK poses difficulties for beginners. Compared to more established frameworks, it has less resources and a smaller community because of its relatively recent inception. There can be a high learning curve to grasp CNTK, which limits its accessibility for individuals with little to no prior deep learning knowledge or for short development cycles. Even though CNTK has strong neural network tuning capabilities, reaching optimal performance can be challenging and needs a thorough understanding of the underlying workings of the framework. It may be necessary for users to do extensive performance tuning, which might be intimidating for those without advanced training or experience.

## 2.4 Torch/PyTorch

Torch is a DL framework written in the Lua programming language. A tensor or an array is used in many of the functions in the torch. These functions include memory exchange, indexing, slicing, and resizing [101]. The torch was designed by Facebook in 2017 and employs dynamic graphics to manage variable-length deliverables. PyTorch is the Python version of Torch [102]. Both PyTorch and Torch depend on dynamic computational graphs and tensors. The fundamental operation entails a linear transformation [103] that is subsequently followed by the activation function:

$$z = Wx + b \tag{4}$$

where $x$ and $z$ represent the input and output tensor, respectively; $W$ denotes the weight tensor; $b$ signifies the bias tensor.

PyTorch has gained considerable acceptance by providing a dynamic computation graph that streamlines imperative programming and permits users to dynamically modify models at runtime. It is written in C, Python, and CUDA and features acceleration libraries from Intel and NVIDIA. This feature has aided PyTorch's rapid growth and adoption among academic communities. By integrating seamlessly with well-known libraries and tools, PyTorch fosters a thriving ecosystem. The machine learning community has witnessed a significant increase in adoption due to its user-friendly API, robust community backing, and seamless transition capabilities from research to production. PyTorch has significantly influenced the development of DL frameworks, thereby bolstering the prevalence of dynamic graph-based methodologies.

Torch has several drawbacks, such as restricted scalability. Thus, it is best suited for smaller applications. PyTorch has come a long way in addressing scalability in production and distributed computing settings. However, it may still be difficult for some applications that demand large-scale deployment and substantial scalability to operate at peak efficiency. The Lua programming language, on which the original Torch framework was based, is becoming less widely used and popular. It is more difficult to locate resources, community support, and knowledge due to Lua's declining popularity in the programming community. This restriction may make development more difficult.

## 2.5 Theano

Theano is a Python deep learning framework that acts as a compiler for numerical expressions and allows developers and users to evaluate their mathematical process via NumPy's syntax [104]. This package was created at the lab of the University of Montreal. Some packages have been designed to increase the capabilities of Theano, such as Pylearn2 [105], Blocks, Lasagne [106], and Keras [107]. This framework facilitates the optimization, effective definition, and calculation of mathematical expressions that incorporate multi-dimensional arrays. Compiling mathematical expressions into highly optimized code, frequently executed on GPUs to enhance performance, is made possible by Theano's symbolic computation approach. Although Theano's development team actively developed and expanded its infrastructure, its development ended in 2017. Theano is a framework for symbolic computation that uses symbolic definitions of mathematical expressions [108]. Matrix multiplication is a frequently performed operation:

$$Z = dot(W, X) + b \tag{5}$$

In Eq. (5), $X$ and $Z$ represent the input and output tensor, respectively; $W$ denotes the weight matrix; $b$ defines the bias vector.

Since Theano is no longer being actively developed, it is not receiving updates or enhancements to take advantage of new developments in deep learning techniques and technology. Its applicability for cutting-edge research and applications that demand the latest features and optimizations is severely limited by this lack of continuous maintenance and development. Theano operates at a lower abstraction level compared to the modern framework. When building deep learning models, its low-level API might be challenging, making it less desirable than more recent frameworks with higher-level abstractions.

## 2.6 TensorFlow

TensorFlow is a deep learning framework that employs an individual data flow graph to define all mathematical operations to provide exceptional throughput [109]. It is an open-source framework that was developed by the Google brain team. TensorFlow is distinguished by its static computation graph, which facilitates the optimization and efficient execution of complicated neural network architectures. Dataflow graphs are used by TensorFlow, with operations as nodes and tensors flowing along their edges [110]. A linear transformation and a non-linear activation are involved in the main operations:

$$y = f(Wx + b) \tag{6}$$

where $x$ and $y$ represent the input and output tensor, respectively; $W$ and $b$ define the weight and bias tensor, respectively; $f(\cdot)$ denotes activation function.

TensorFlow creates huge computational networks wherein every node in the graph refers to a mathematical function, and the edges indicate node interaction. This data flow graph specifically conducts interaction between a large part of the computational system, allowing for concurrent execution of separate calculations or the deployment of numerous devices to run partition operations [111]. TensorFlow Lite (enable low-latency inferences), TensorFlow serving (high performance, open source serving system), TensorFlow Hub (library for the publication, discovery, and consumption), and TensorFlow.JS are closely associated with or variants of TensorFlow (build and train models entirely in the browser) [112]. A general framework of TensorFlow components is illustrated in **Fig. 2(a)**. TensorFlow constructs and executes these graphs using APIs from multiple programming languages, including Python, C++, and Java. TensorFlow continues to develop, incorporating features such as TensorFlow Extended (TFX) **(Fig. 2(b))** to facilitate end-to-end machine learning workflows and TensorFlow Lite to enable model deployment on mobile and edge devices, all in support of a large and active community.

The original low-level API for TensorFlow required users to handle variables, maintain sessions, and create computational networks manually. Creating models can be laborious and time-consuming due to this degree of detail, especially for users who are used to higher-level, more intuitive interfaces. The development process becomes more complicated when computation graphs must be manually created and run, frequently leading to lengthy and complex code. This low-level control might be a major obstacle for novices or those wishing to rapidly prototype models. Furthermore, despite its adaptability, TensorFlow's complexity may be challenging for consumers who want ease of use or quick development.

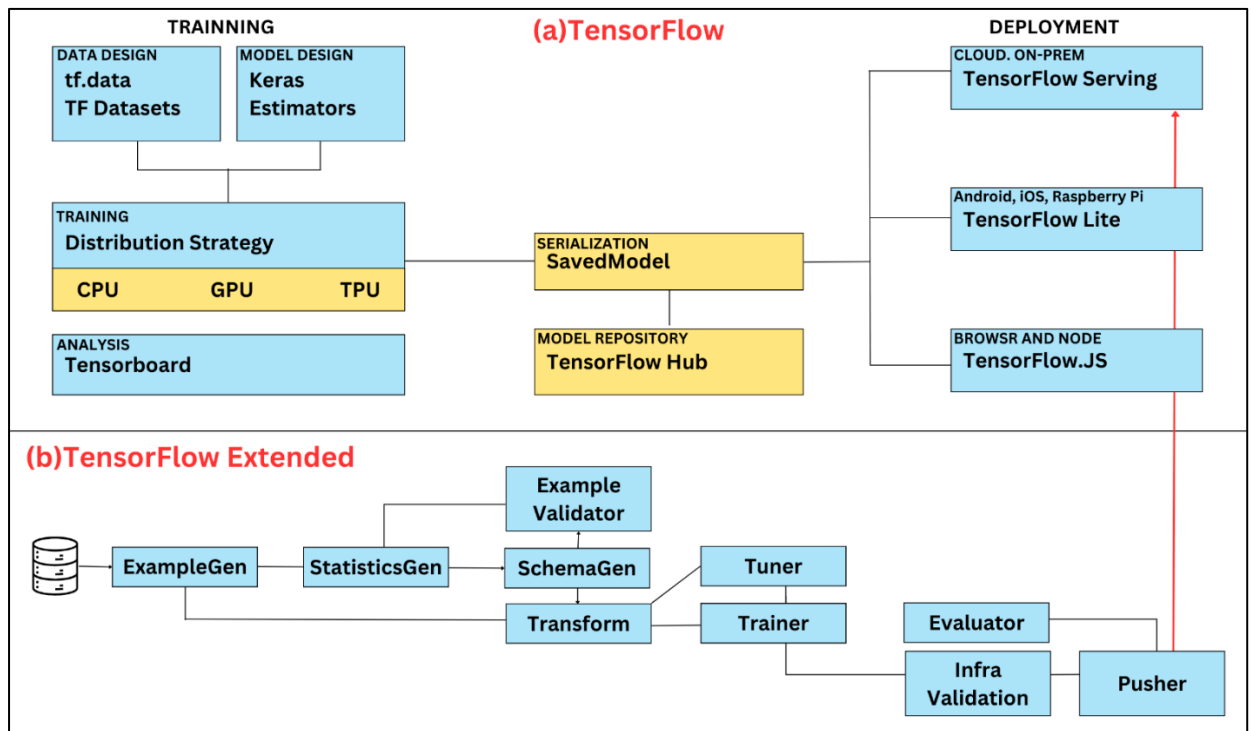

**Fig. 2. (a)** TensorFlow pipeline and key components; **(b)** TensorFlow extended pipeline.

## 2.7 Chainer

Chainer is an open-source framework for deep learning distinguished by its adaptability and dynamic construction of computational graphs [83]. Chainer, developed by Preferred Networks, became well-known for its innovative methodology allowing users to dynamically define and alter neural network architectures while the program ran. In contrast to static computation graphs utilized by certain alternative frameworks, this dynamic approach offers practitioners and researchers considerable adaptability to facilitate experimentation and fast prototyping. With Chainer's "define-by-run" technique, computational graphs are built dynamically while the program executes [83]. An activation function is typically implemented subsequent to a linear transformation:

$$y = \sigma(Wx + b) \tag{7}$$

where $x$ and $y$ represent the input and output tensor, respectively; $W$ and $b$ define the weight and bias tensor, respectively; $\sigma(\cdot)$ denotes activation function.

Chainer is applicable to many applications due to its support for both CPU and GPU acceleration. A variety of DL models are offered by Chainer, such as reinforcement learning (RL), variational autoencoders, CNN, and RNN [91]. Chainer, renowned for its intuitive interface and imperative programming compatibility, has garnered extensive implementation across a spectrum of machine learning applications, encompassing natural language processing and image analysis. It has proven to be a valuable instrument for researchers investigating innovative architectures and algorithms within the ever-evolving domain of deep learning due to its user-friendly interface and dynamic graph construction.

The comparatively smaller community and ecosystem of Chainer in comparison to other deep learning frameworks such as TensorFlow or PyTorch is one of its primary limitations. Due to this restriction, users may have less access to help resources, including tutorials, discussion boards, and third-party libraries. Compared to extensively used frameworks, Chainer is rarely used in commercial and industrial environments. Its low industry adoption may lessen its appeal to companies seeking industry-standard tools. Enterprise applications may be better served by frameworks with a longer history and more community support.

## 2.8 PaddlePaddle

An open-source DL framework developed by Baidu [113], PaddlePaddle is also referred to as Paddle. PaddlePaddle is a highly adaptable framework specifically engineered to encompass various uses, including linguistic recognition, reinforcement learning, computer vision, and more. Its emphasis on supporting static and dynamic computation graphs is one of its distinguishing characteristics; this allows users to optimize and design neural network models with greater flexibility. A non-linear activation followed by a linear transformation is the basic operation [113]:

$$y = ReLU(Wx + b) \tag{8}$$

In Eq. (8), $x$ and $y$ represent the input and output tensor, respectively; $W$ and $b$ define the weight and bias tensor, respectively; $ReLU(\cdot)$ denotes the activation function.

PaddlePaddle architecture comprises four components: core framework, visual interface, scene suites, and algorithm suites [85]. The outstanding suitability of PaddlePaddle for large-scale and distributed training makes it a practical option for industry applications. The framework also provides users with an extensive collection of pre-built models and tools, which facilitate the process of development. PaddlePaddle's dedication to providing an intuitive and readily available interface has significantly improved its prominence, particularly within the Chinese deep learning community. In general, PaddlePaddle acts as a resilient and entire educational and research platform that facilitates the investigation and execution of state-of-the-art deep learning methodologies.

PaddlePaddle's limited acceptance outside of particular locations, especially given its strong correlation with Chinese technology ecosystems. Its use and visibility across the international development community may be limited by this geographical specialization. Because of this, developers and organizations in areas where alternative frameworks like TensorFlow or PyTorch are more popular might not choose it initially. In contrast to other widely used frameworks, this may limit its community support and the availability of pre-built models or tools.

## 2.9 Keras

Keras is a renowned open-source framework for deep learning due to its user-friendly and simple architecture [91]. Keras, originally designed to facilitate the construction of neural networks from other deep learning frameworks (e.g., TensorFlow), has undergone significant development and is now a vital component of the TensorFlow ecosystem. Its high-level API enables developers to rapidly prototype and experiment with neural network architectures, making it suitable for beginners and professionals [114]. The primary operation of Keras is characterized by layer-based abstractions [115]:

$$y = f(Wx + b) \tag{9}$$

where $x$ and $y$ represent the input and output tensor, respectively; $W$ and $b$ denote the weight and bias tensor, respectively; $f(\cdot)$ denotes activation function.

By removing numerous complications associated with lower-level implementations, Keras offers an efficient and user-friendly interface that facilitates developing, training, and deploying deep learning models. The seamless integration with diverse backends, such as TensorFlow and CNTK, is facilitated by its flexible nature. Keras allows users to concentrate on the design and structure of their neural networks without getting involved in complex implementation details [116], leading to its extensive use in research and industry applications. This characteristic has enabled Keras to gain extensive acceptance in both academic and industrial sectors.

TensorFlow and Keras are generally interwoven, and although this connection has numerous advantages, there are drawbacks as well. Because of this close relationship, Keras users frequently use TensorFlow for backend processing. Because of this, integrating Keras with other deep learning frameworks may be difficult or time-consuming. For simplicity and quick development, Keras provides a high-level interface by abstracting many of the underlying intricacies of model building and training. For users who want precise control over their models, this abstraction may be a drawback. Due to its higher-level nature, tasks requiring precise tuning, the generation of custom layers, or certain optimization techniques could be difficult to accomplish in Keras.

***Summary***: Contemporary artificial intelligence and machine learning applications are supported by deep learning frameworks, which offer programmers powerful tools for the construction, training, and implementation of neural network models. Caffe2, which is adaptable and mobile-compatible, enhances the efficiency of its predecessor, Caffe, which was established on the foundation of speed. MXNet distinguishes itself through its exceptional scalability and compatibility with multiple languages, whereas CNTK prioritizes peak performance and effective training. PyTorch and Torch, its successors, provide simple interfaces and dynamic computation graphs, supporting a diverse group of developers. Formerly an innovator in symbolic expressions, Theano's popularity has recently declined to be replaced by simpler alternatives. Chainer prioritizes adaptability, PaddlePaddle simplifies model construction, and Keras, now integrated with TensorFlow, emphasizes usability. TensorFlow is a foundation in the deep learning domain due to its extensive usage, scalability, and adaptability. Every framework possesses unique advantages and disadvantages, as outlined in Table 2, that impact its appropriateness for particular use cases and user choices. These frameworks collectively contribute to expanding and progressing deep learning technologies as the field develops. Table 3 explicitly represents the potential advantages of these frameworks.

Deep learning frameworks offer indispensable libraries and tools that enable developers and researchers to efficiently experiment with and implement deep learning algorithms, promoting innovation and breakthroughs in different disciplines. In order to evaluate the performance of deep learning frameworks across a variety of applications, it is necessary to take into account the particular requirements of each field. For instance, renowned for their adaptability and backing from the community, TensorFlow and PyTorch perform admirably in natural language processing, computer vision, and bioinformatics. Agriculture may benefit from PaddlePaddle's emphasis on usability. Due to their robust libraries, TensorFlow, PyTorch, and Keras dominate drug discovery and financial fraud detection. Fluid dynamics simulations are frequently conducted using TensorFlow or PyTorch, whereas TensorFlow's serving capabilities render it well-suited for real-time endeavors within disaster management.

Table 2. Advantages and disadvantages of the DL frameworks, with brief descriptions of each

| DL frameworks | Brief description | Advantages | Disadvantages |
|---|---|---|---|
| Caffe | Caffe represents convolutional architecture for rapid feature embedding. It is a freely available DL framework that may be used to analyze complicated networks. | – The code is ideally suited for research due to its flexibility.<br>– Caffe nicely functions with CNN architecture.<br>– It allows the adjustment of training models without writing perfect code. | – Caffe does not have good documentation.<br>– It performs poorly and is cumbersome in big neural networks and RNN architecture.<br>– Its development has become slow. |
| Caffe2 | Caffe2 is an improved variant of Caffe. It fixes a number of issues with Caffe, including scalability, portability, and the lack of hardware support for quantized processing. | – Caffe2 is a cross-platform deep learning framework. It can be used in mobile networking systems or edge computing systems.<br>– Caffe2 is said to be supported by Amazon, Intel, Qualcomm, and NVIDIA because of its scalability and robustness. | – Compatibility with Caffe2 can be challenging for new users.<br>– It does not support dynamic graph computations. |
| MXNet | MXNet is a fast and flexible deep learning framework developed for several languages. It has a small memory footprint while still being extremely scalable. | – It supports high-level programming languages such as C++, R, Python, Scala, JavaScript, Perl, Go, Julia, Matlab, and Wolfram.<br>– It supports dynamic graph computations.<br>– Open Neural Network Exchange (ONNX) format also works in MXNet, making switching between other DL frameworks and libraries simple. | – It is not compatible with some APIs. |
| CNTK | As a deep learning framework, CNTK is written in C++ and supports a Python API. CNTK simplifies combining popular network architectures, including CNNs, Feed-forward DNNs, and RNNs/LSTMs. | – CNTK provides dependable and superior functionality.<br>– It's compatible with Azure Cloud, which is supported by Microsoft.<br>– The control and use of resources are both efficient. | – It is hard for beginners.<br>– As it's significantly new, it does not have vast community support. |

| Torch/PyTorch | The Torch framework for deep learning was developed in Lua.<br>Many Torch operations depend on either a tensor or an array. | – Torch contains a lot of effective extensions and support<br>– It provides GPU support that is both quick and effective.<br>– It is research-friendly. | – Only the smaller scale projects can be run on this<br>– Lua has become a mid-level programming language; thus, it is not popular anymore. |
|---|---|---|---|
| Theano | Theano is a numerical expression compiler for the Python deep learning framework NumPy, enabling both programmers and data scientists to assess the efficacy of their mathematical procedures in the context of the framework. | – Theano is a cross-platform framework and open source.<br>– It provides support in GPU, and mathematical calculations can be done faster.<br>– It makes RNNs efficient because of symbolic API. | – No longer under development.<br>– Low-level API.<br>– Theano is challenging to utilize explicitly for constructing deep learning models. |
| TensorFlow | The TensorFlow deep learning system uses a single data flow graph to express all mathematical operations, resulting in extremely high throughput. The Google Brain team developed this open-source platform to help advance artificial intelligence. | – TensorFlow is an open-source, popular, and promptly changing available framework.<br>– It has a powerful numerical library as the foundation for deep learning research.<br>– Efficiently work with numerical expressions which have a multidimensional array.<br>– It has great computational versatility across machines and large datasets. | – A low-level API makes generating DL models tough. |
| Chainer | Chainer is an open-source deep learning framework characterized by its dynamic construction of computational graphs and adaptability. | – Chainer's performance for deep learning tasks is improved by its seamless integration with GPUs for rapid computation.<br>– It enables flexibility in model development and runtime modification.<br>– The intuitive interface of Chainer renders it suitable for users of all skill levels, including beginners. | – Although Chainer experiences an active user community, it may lack resources and community support compared to more widely adopted frameworks.<br>– In comparison to other DL frameworks, its utilization in industry may be comparatively infrequent. |

| PaddlePaddle | PaddlePaddle stands out for its support for both static and dynamic computation graphs. This gives users more flexibility when optimizing and designing neural network models. | – PaddlePaddle is well-suited for managing complex models and large datasets because it supports large-scale and distributed training.<br>– It simplifies the development process through its user-friendly interface, comprehensive tools, and pre-built models. | – Its adoption may be comparatively limited in scope and recognition beyond specific regions. |
| --- | --- | --- | --- |
| Kera | Keras is widely recognized as an open-source DL framework due to its user-friendly and simple architecture. It was designed to facilitate the construction of neural networks from other deep learning frameworks, such as TensorFlow. | – Keras offers a high-level API that simplifies intricate operations, thereby facilitating the development and training of neural networks by users with minimal coding.<br>– Keras's compatibility with multiple backends, including TensorFlow and Microsoft Cognitive Toolkit, provides users with adaptability and the ability to operate in diverse ecosystems. | – Although Keras demonstrates adaptability, its integration with alternative deep learning platforms may be hindered by its close association with the TensorFlow ecosystem.<br>– It may not provide the same degree of low-level control and flexibility for advanced research as alternative frameworks. |

Table 3: Comparative analysis of various deep learning frameworks listing potential advantages

| Potential Advantages | Caffe | Caffe2 | MXNet | CNTK | Torch/PyTorch | Theano | TensorFlow | Chainer | PaddlePaddle | Keras |
|---|---|---|---|---|---|---|---|---|---|---|
| Flexibility for research | √ | √ | √ | √ | √ | × | √ | √ | √ | √ |
| Ease of use for beginners | × | × | × | × | √ | × | × | √ | √ | √ |
| Supports dynamic graph computation | × | × | √ | √ | √ | × | √ | √ | √ | √ |
| Good documentation and community support | × | × | √ | × | √ | × | √ | × | × | √ |
| Scalability and distributed training | × | √ | √ | √ | √ | × | √ | × | √ | × |
| Cross-platform compatibility | √ | √ | √ | √ | √ | √ | √ | √ | √ | √ |
| High-level API availability | × | × | √ | × | √ | × | × | √ | √ | √ |

×: Not available; √: Available

## 3. Recent advancements and applications of deep learning

Deep learning has addressed challenges that were considered to be impossible a few years back. Specifically, it has successfully handled issues that conventional machine algorithms failed to address; for example, deep learning has revolutionized image and video analysis, where traditional methods often struggle with variations in lighting, angles, and backgrounds. Because of its exceptional data handling feature, it has piqued the interest of professionals who are bombarded with all kinds of data. The advancement and application of deep learning (**Fig. 3**) have increased.

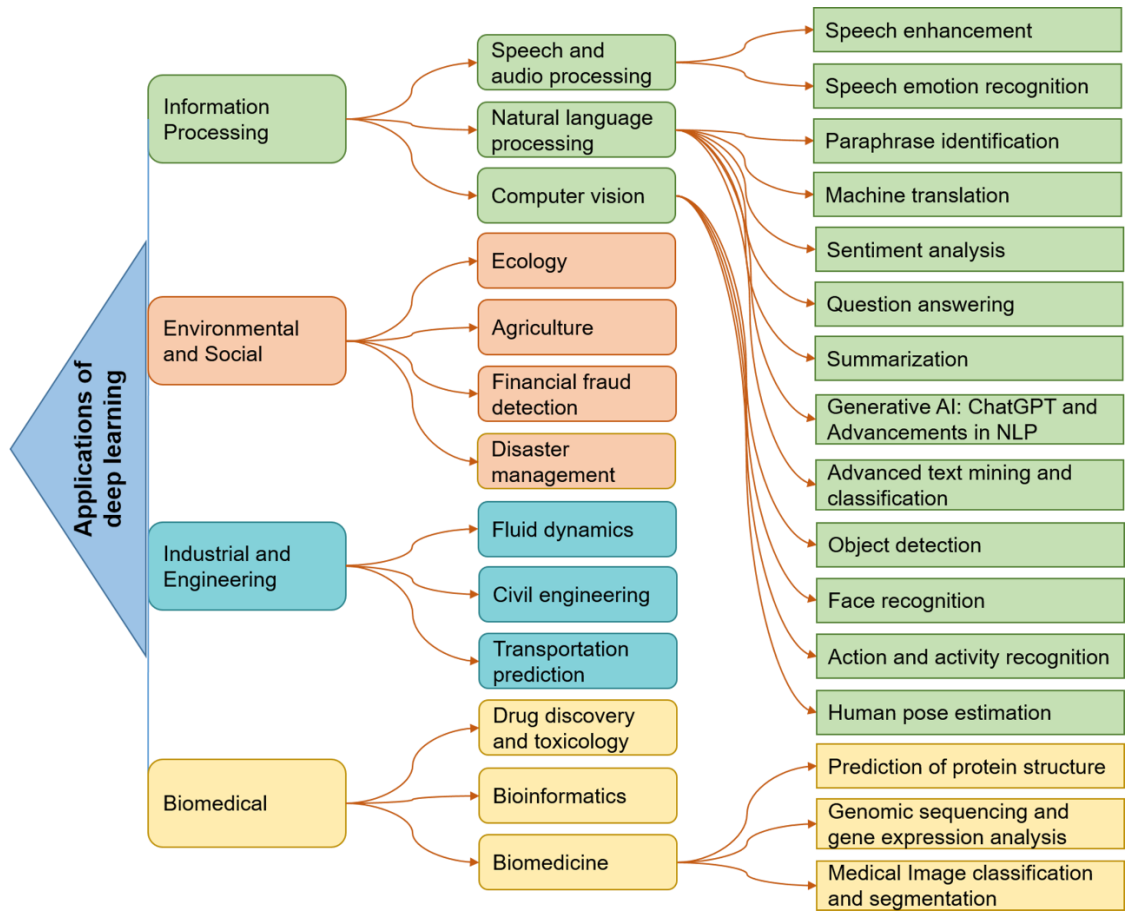

**Fig. 3.** Applications of deep learning in potential sectors

Natural language processing is one of the fields that has benefited significantly from deep learning, which is used in the various domains of natural language processing, like the translation of audio and machines [117]. In 2015, Google debuted the word lens identification engine based on deep learning. The best feature of a word lens is that it can read the texts instantaneously and convert them into the target language [118]. Another field where deep learning is being immensely used is transportation. As the world population increases, a smart transportation system is in demand. In the transportation sector, deep learning is being used in Destination Prediction [119], Demand Prediction [117,118], Traffic Flow Prediction [122–124], Travel Time Estimation [125, 126], Predicting Traffic Accident Severity [127, 128], Predicting the Mode of Transportation [129], Trajectory Clustering [130, 131], Navigation [132, 133], Demand Serving [134, 135], and Traffic Signal Control [136, 137]. Implementation of DL in the bioinformatics domain has increased substantially over the years. In this area, DL has been mostly used to predict the structure of proteins [138, 139], gene regulations and expressions [140, 141], precision medicine[142, 143], biomedical imaging [144, 145], localization of cells [146, 147], clustering [27, 148], classification of proteins [149, 150] and so on. Techniques of DL have become one of the key elements in many multimedia systems [151]. Systems like Convolutional Neural Networks (CNNs) have presented noteworthy outcomes in various tasks from the real world, such as the detection of objects and processing of visual data, including images and videos. There have been numerous advancements, including the development of event detection from sports videos [152], new techniques like recurrent convolution networks (RCNs) for video processing[153], the use of intermediate CNN layers, and Gated Recurrent Units for improving the sparsity and locality in modules.

### 3.1 Information processing

#### 3.1.1 Speech and audio processing

Speech and audio have been proven to be two important modes of communication in human history. Deep learning technologies have made significant progress in the field of speech processing. Unlike traditional speech enhancement techniques that rely on a statistical model, deep learning models are data-driven. Deep learning is among the most important technologies nowadays and, thus, merits its study. Mapping-based and masking-based are DL approaches used in speech enhancement [154]. Moreover, DL methods for speech emotion recognition have several advantages over conventional machine learning methods, such as identifying complex systems and no requirement for manual extraction of the features [155]. Also, deep learning has the ability to handle a large amount of unlabeled data and has a proclivity for obtaining low-level characteristics from provided raw data. Furthermore, many academics have abandoned traditional signal-processing approaches for sound production due to the emergence of deep learning algorithms. Deep learning techniques have accomplished eloquent voice generation, audio textures, and melodies from simulated instruments [156]. Deep neural networks have shown excellent progress in audio processing.

#### 3.1.1.1 Speech enhancement

Speech enhancement is utilized in video conferencing, hearing aids, microphones, audiovisual screening, and so on. In order to have a good speech enhancement system, pattern mining is needed [157]. Processing background noise is considered a good speech augmentation method. There are different types of conventional machine learning approaches available to filter and eliminate the additional noise from the voice signals. In recent years, DL methods have proven beneficial for speech augmentation. Deep neural networks (DNNs) can learn and approximate complex patterns in the data due to the nonlinearity introduced by activation functions. The model's ability to generalize, manage neuron activation levels, and adjust to the particular demands of various tasks and architectures is further impacted by the activation function selected. Many activation functions may be employed in deep neural networks for speech enhancement; the selection often relies on the particular demands of the required tasks and the nature of the data. Speech enhancement applications frequently employ the exponential linear unit, sigmoid and hyperbolic tangent, GRU and LSTM, and rectified linear unit as activation functions.

The DNN is the most general technique to remove noises from large datasets [158]. However, research shows that DNN struggles to adapt to new voice corpora in low signal-to-noise ratio (SNR) settings [159]. Accordingly, an alternative solution is to utilize a lower frameshift in short-time speech processing that can substantially improve cross-corpus generalization. Noise suppression is a well-developed field of signal processing [160], yet it still heavily relies on the precise tweaking of estimator techniques and variables. In a study by Valin [161], hybrid deep learning technologies were proposed to eliminate noise from the background. In this technique, a great emphasis is placed on minimizing complications while obtaining highly improved speech. This method proves more viable than a standard lowest mean squared error spectrum estimator.

In order to achieve a significant speech enhancement performance, a large DNN is needed, which is computer- and memory-intensive [162]. Therefore, making such a speech enhancement system is challenging because of timing constraints and minimal hardware resources. One study introduced two compression models for DNN-based speech enhancements, mainly including quantization to minimize model size based on clustering, sparse regularization, and repeated pruning [163]. Moreover, experimental results indicate that this method decreases the size of four distinct models by a substantial amount without affecting their enhancing performance.

The DNN model was also used by Mukhutdinov et al. [164] to evaluate the efficiency of speech enhancement by reducing drone ego-noise. Twelve illustrative DNN models were trained, spanning three different architectures (encoder-decoder, sequential, and generative) and three different operation domains (end-to-end time, time-frequency complex, and time-frequency magnitude). In scenarios with extremely low signal-to-noise ratios (SNRs), ranging from -30 dB to 0 dB, the models' performance was assessed. While competing well on measures of speech enhancement, time-frequency complex domain and UNet encoder-decoder offered a reasonable compromise with respect to other metrics, including model size, context length, and computational complexity. The UNet model performed the best in the time-frequency complex domain and was able to increase ESTOI from 0.1 to 0.4, PESQ from 1.0 to 1.9, and SI-SDR from −15 dB to 3.7 dB, all with an input SNR of −15 dB.

### 3.1.1.2 Speech emotion recognition

Speech emotion recognition (SER) refers to identifying the underlying emotion in a text or speech regardless of the semantic content. Implementation of deep learning in SER has made it possible to detect emotion in a speech in real-time situations [165]. Emotions can generally be categorized and organized into different groups, such as happiness, anger, sadness, and even a neutral emotional mood. Several researches have aimed to enhance the SER technique using deep learning methods. In a study, a dynamic SER identification system was developed based on aural and visual data processing [166]. In this system, a total of 88 characteristics (Mel Frequency Cepstral Coefficients (MFCC), filter bank energies (FBEs)) are employed to minimize the earlier retrieved features using principal component analysis (PCA) [167]. A comparison of deep learning flow versus conventional machine learning flow techniques for SER is shown in **Fig. 4**.

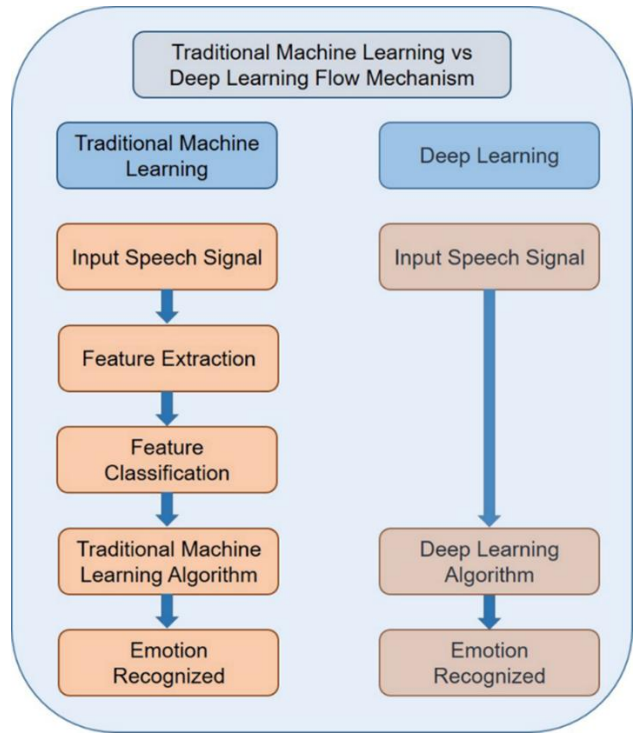

**Fig. 4.** Traditional machine learning vs. deep learning flow process (modified from [168])

Surekha [169] utilized Gaussian mixture model (GMM) classifiers in the SER system to detect five emotions by applying the Berlin Emotional Speech database. An acoustic feature called the MFCC with the Teager Energy Operator (TEO) was used as a prosodic. This approach was employed in another study where 13 MFCC special features derived from the audio data were utilized to identify seven emotions [170]. In this method, the Logistic Model Tree (LMT) method was used, which has a 70% accuracy rate. Nevertheless, there are some challenges related to SER that need to be addressed. For instance, an issue with emotional speech datasets is annotation ambiguity [171]. In a specific task like picture classification, a car will always be labeled as a car, yet in an emotional discourse, asserting something strongly may be categorized as anger. This bias in categorization complicates the work while limiting the ability to combine datasets and generate emotional supersets.

A study by Hema et al. [172] introduced a SER system that demonstrated superior performance compared to an earlier system in terms of data collection and feature selection techniques. The study introduced the concept of SER and looked at how emotions play an important role in human communications. As mentioned earlier, this technology analyzes audio signals to extract and predict the emotional state of a speaker. The SER task utilized support vector machine (SVM), radial-basis function (RBF), and back propagation ML models. These sophisticated models outperformed current systems' performance in collecting data and selecting feature techniques, resulting in an accuracy rate of 78% with a reduced number of false positives. Feature normalization can mitigate the impact of speakers' diversity on recognition.

### 3.1.2 Natural language processing

Deep learning algorithms are rapidly being implemented in natural language processing (NLP) research. Popular ML algorithms, such as SVM and logistic regression, trained on sparse and high-dimensional features have been the foundation of ML approaches for NLP issues for decades. On a range of NLP tasks, neural networks based on dense vector representations have recently surpassed classical models. This trend has been fueled by significant advancements in word embedding and DL approaches [117]. Classical NLP relies on predefined features and protocols, while deep learning NLP trains directly from data representation, which makes it more adaptable and capable of handling complex tasks. The abstract framework of both classical and DL-based frameworks is demonstrated in **Fig. 5**.

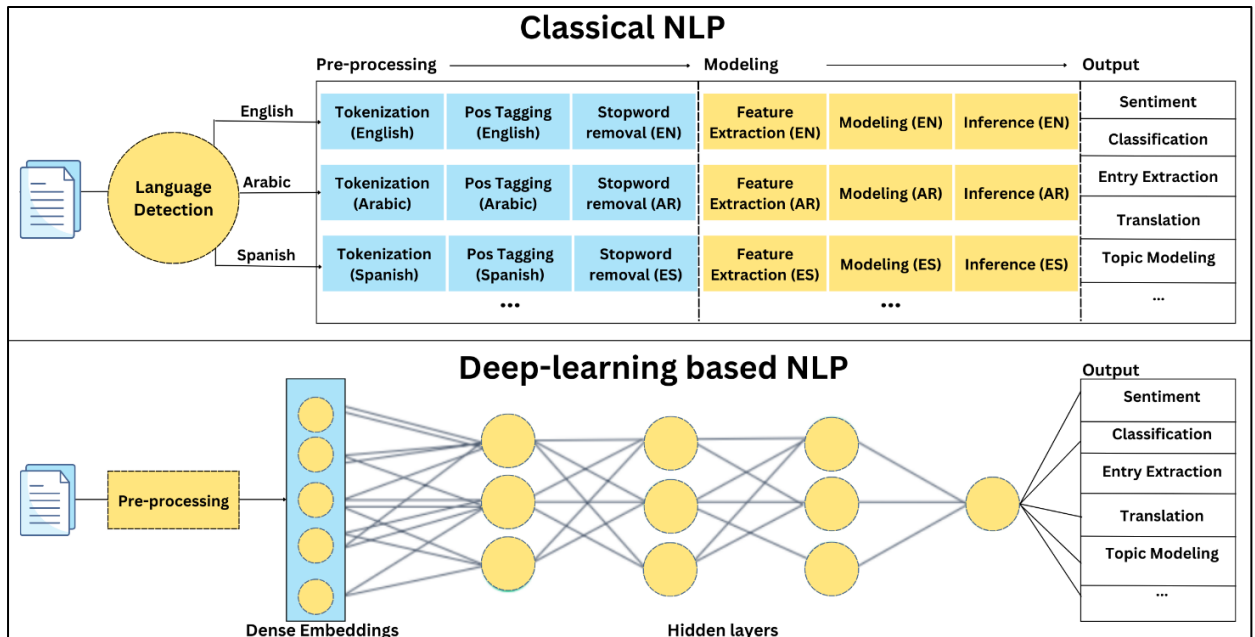

**Fig. 5**. Framework of classical NLP and DL-based NLP.

Word embedding models based on Bidirectional Encoder Representations from Transformers (BERT) have shown effective for several NLP tasks. For instance, three BERT implications, such as regular, ClinicalBERT, and BioBERT, were performed in a study by Turchin et al. [173]. The study evaluated the accuracy of three medical concepts that are linguistically complex. These concepts were i) patients and healthcare providers discussing bariatric surgery, ii) patients not accepting statin treatment recommendations, and iii) documentation of tobacco use status. Compared to the regular BERT trained exclusively on Wikipedia, ClinicalBERT and BioBERT achieved better NLP performance for identifying various complicated medical concepts. This is due to the fact that they are trained on documents from the biomedical domain. The two biomedical BERT implementations tested did not achieve a significantly higher level of accuracy than the other.

### 3.1.2.1 *Paraphrase identification*

Identifying whether two statements written in natural language have comparable semantic meanings is known as identification. When two sentences have the same meaning, they are referred to as paraphrases. This methodology is a fundamental approach in various data mining approaches that has tremendous potential in various domains, including plagiarism detection, machine translation, and others. Various techniques of paraphrase identification have been presented, which are distributed into two categories: similarity-based and classification-based methods [174]. Alzubi et al. [175] applied a collaborative adversarial network (CAN) in order to discover the problem of paraphrasing. A common feature extractor is included in CAN to help increase the association between phrases in the recurrent neural network model. Within a word pair, the extractor looks for related features. The integration of adversarial networks with collaborative learning mostly enhances the generator and discriminator working together. The model outperforms the baseline MaLSTM model as well as several baseline approaches. Hunt et al. [176] investigated several ML approaches for modeling the problem of paraphrase identification and alternative input encoding schemes. Since RNN outperformed the other models, the researchers developed a method based on the LSTM, an RNN variation. RNN works by using the output of past time sequences to create the output for next time sequences. This fits in NLP because it can be illustrated as a sequence of tokens, and the appearance of the next token is generally influenced by the preceding tokens.

### 3.1.2.2 Machine translation

NLP's most well-known application is machine translation, which entails translating texts from one language to another using mathematical and computational procedures. The translation is challenging for humans since it requires knowledge of syntax, semantics, and morphology, as well as a full understanding and evaluation of cultural sensitivity for the target language and its associated communities. Ahmed et al. [177] advocated that attention methods can be used to encode a language from input to output instead of employing a huge number of recurrent and convolutional layers. The following three principles motivate the use of "self-attention" methods over traditional layers: decreasing the complexity of calculations needed for each layer; limiting sequential training steps; and lastly, reducing the length of the path between input and output and its influence on the learning of long-range relationships, which is important in several sequencing tasks.

Johnson et al. [177] reported that a single, basic (but huge) neural network could be applied to translating a number (minimum 12) of several languages into each other. This neural network automatically recognizes the source languages and uses just one input token to determine the output languages. When numerous language tokens are supplied, the method is shown to be capable of interpreting multilingual inputs and producing mixed outputs, at least in part, sometimes even in languages similar to but not identical to those chosen. The performance of such zero-shot translation is frequently insufficient to be practical, as the basic pivoting strategy quickly outperforms it.

A deep-attention technique was introduced by Zhang et al. [178] for a neural machine translation (NMT) system. This model incorporates many stacked attention layers, each of which pays attention to a corresponding encoder layer to advise what should be transmitted or repressed from the encoder layer, resulting in learned distributed representations that are suitable for high-level translation tasks. English-French, NIST-English-Chinese, and WMT14

English-German translation skims were used in this experiment. The technique could be implemented on other tasks like summarization and adapted to more complex attention models. Aside from the work listed above, other academics have suggested a variety of high-performance designs. In the English-German and Chinese-English translation systems, Zhang et al. [179] introduced the Variational NMT approach, which has a novel approach to modeling translation problems. The results demonstrated that it performed better than the baseline technique. A fast-forward connection for RNN was developed by Zhou et al. [180], which allows for a deeper network in implementation and, hence, greater performance.

### 3.1.2.3 Sentiment analysis

Users today create massive volumes of data largely and dynamically as the number of internet technologies grows. The number of people who use social media on a regular basis is continuously expanding. In this sense, sentiment analysis appears to be a useful technique for automating the extraction of insights from user-generated data. Deep learning algorithms have recently been presented for several sentiment analysis applications, with the existing results [181]. A deep learning-based sentiment analysis pipeline is depicted in **Fig. 6**. Text input, such as a review or tweet, is analyzed by a neural network equipped with an attention mechanism. The output is a sentiment distribution displayed as a bar graph, indicating probabilities for emotions such as positive, neutral, and negative. The pipeline demonstrates how deep learning models process textual data to derive meaningful insights.

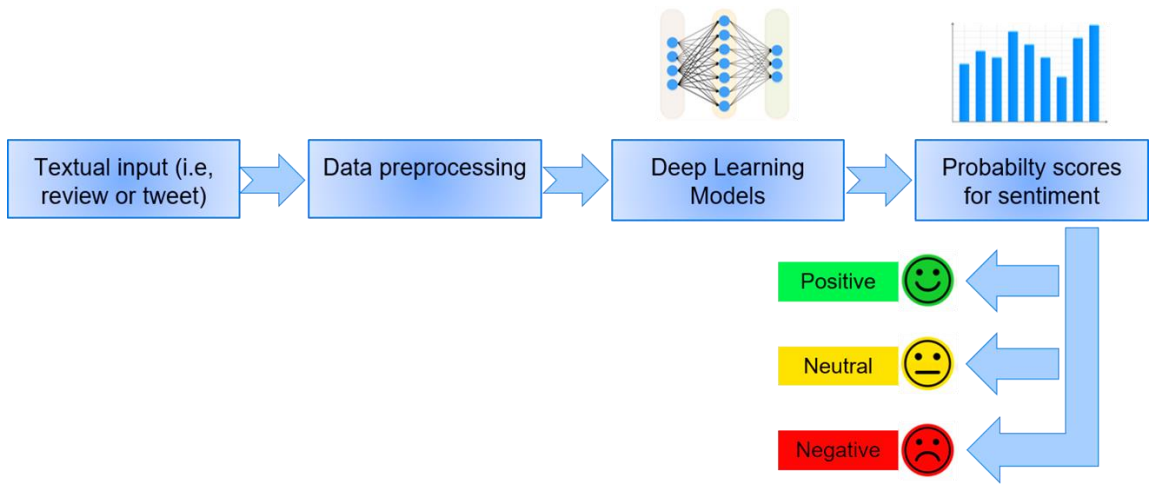

**Fig. 6.** Deep learning-based sentiment analysis pipeline

For extracting aspect opinion target expressions (OTEs), Al-Smadi et al. [182] suggested combining BiLSTM with a CRF method for classifying, as well as for classifying aspect-sentiment polarity. An aspect-based LSTM was used, where the aspect opinion target expressions were applied as attention expressions. A two-stage attention structure was created by Ma et al. [183] that involves paying attention to both the words that make up the target expression and the complete phrase. The author also used extended LSTM to construct a common sense model for target aspect-based emotion detection, which can use external knowledge. However, these methods could not represent several aspects of sentences, and the explicit location contexts of words were not investigated.

A path aggregation relational network for aspect-based sentiment analysis (ABSA) was proposed by Ma et al. [184] that is based on abstract meaning representation (AMR). This model used the AMR semantic structure, which is better in line with the sentiment analysis problem, as opposed to the old ABSA model, which used syntactic structures like dependency trees. The model utilized BERT, a well-known transformer-based model, to encode sentences. Specifically, the authors designed the relation-enhanced self-attention mechanism and the route aggregator, which work well together. Sentence information guides the route aggregator in extracting semantic features from AMRs, and the relation-enhanced self-attention mechanism refines sentence characteristics using this extracted semantic information. When tested against state-of-the-art baselines, experiments on four open data sets showed that AMR-based path aggregation relational network (APARN) improves ABSA by an average of 1.13%. However, the level of accuracy of the AMR parsing results continues to slightly impact the model's performance.

Sentiment analysis has become extensively utilized in microblogging platforms like Twitter in the past decades due to the concise and simple nature of expression, allowing millions of users to share their opinions and thoughts. The BERT exhibits greater comprehension compared to traditional models due to its ability to process all input data through the encoder. Bello et al. [185] proposed to use BERT for text classification in NLP, along with other different variants. After comparing the results with Word2vec and no variant, the experimental results showed that combining BERT with CNN, RNN, and BiLSTM performed better in terms of accuracy, precision, recall, and F1-score. However, the conventional approach of NLP that employs Word2Vec, CNN, RNN, and BiLSTM possesses a few constraints, including the inability to capture the word's deeper context.

By incorporating diverse modalities (e.g., text, images, and audio), multimodal learning approaches improve sentiment analysis by capturing complex expressions and contextual information. This not only expands the informational scope but also enhances the overall performance of DL models by capturing a wider range of media formats and sentiments. Multimodal learning, employing both shared neurons for general activities and specialized neurons for specific tasks, revolutionizes deep learning models by integrating information from various sources like text, images, and audio [186]. The shared neurons capture overarching patterns and general features common across modalities, fostering knowledge transfer. Concurrently, specialized neurons focus on task-specific nuances, refining the model's understanding of particular objectives. This dual neural architecture enhances the overall effectiveness of deep learning models, providing versatility, adaptability, improved robustness, and generalization.

### *3.1.2.4 Question answering*

Question answering (QA) is the process of extracting words, phrases, or sentences from documents that are relevant. In response to a request, QA logically delivers this information. The approaches are similar to the summarizing methods presented in the next section. Wang et al. [187] utilized an attention-based LSTM to associate the questions with answers containing paragraphs. By mapping the full text, a self-matching attention process was employed to develop the machine representations. The position and boundaries of responses were predicted using pointer networks. The networks utilized attention pooling vector representation of passages, and the words were examined to model the crucial tokens or phrases. Modeling more than one aspect simultaneously with the attention mechanism would be an interesting addition to the experiment. Using Wikipedia as the knowledge source, Mozannar et al. [188] investigated the topic of open domain factual Arabic QA. They proposed an approach composed of A BERT document reader and hierarchical TF-IDF document retriever. They also introduced a dataset (ARCD). However, ARCD's questions were created with certain paragraphs in mind; without that context, they might seem ambiguous.

CNNs were compressed with LSTM to speed up processing [189]. For the decomposition of fully connected layers in LSTM and CNN, the researchers suggested applying different decomposition algorithms and regression techniques. Tensor Regression layers replace the Flattening and Fully Connected layers in the final section of the method. The Tensor Contraction layer compresses the flow of features between the layers to compress the parameter further. Determining the rank is a hard NP problem in low-rank decomposition, and their method is still constrained in this area by inserted hyper-parameters. Yu et al. [190] used a tree-LSTM method for capturing a language's linguistic structures. The study created a semantic tree for each question in the dataset, with every node referring to a single LSTM unit and the root node representing the sequence. To boost reasoning capacity, this approach can divide problems into several logical phrases. The representative ability of the network could be improved in the future.

### 3.1.2.5 Generative AI: ChatGPT and Advancements in NLP

The field of NLP has witnessed transformative developments with the advent of generative AI models such as ChatGPT. Built on the GPT (Generative Pre-trained Transformer) architecture, ChatGPT employs the transformer model's attention mechanisms to understand and generate human-like text. Its training involves two key stages: pretraining on a massive amount of text using autoregressive language modeling and fine-tuning for task-specific applications through reinforcement learning from human feedback [191]. ChatGPT can facilitate cross-lingual communication by providing real-time translation services and addressing language comprehension issues in international interactions [192].

ChatGPT excels in tasks like conversational AI, text summarization, and question answering by leveraging contextual embeddings to maintain coherence over long textual sequences. For instance, compared to earlier architectures such as RNNs and LSTMs, which struggled with long-term dependencies, ChatGPT's self-attention mechanism enables efficient handling of long-range contexts, making it highly effective in dialog systems [191]. However, challenges persist, such as susceptibility to producing plausible but factually incorrect responses, and biases embedded in training data. The widespread adoption of ChatGPT underscores its impact on democratizing NLP tools, yet it also raises ethical concerns related to misinformation (**Fig. 7**), intellectual property, and data privacy. Future research should focus on addressing these challenges and enhancing the transparency and explainability of such models.

| Benefits | | Challenges | |
|---|---|---|---|
| ✓ Scalability<br>✓ Innovation catalyst<br>✓ 24/7 Availability<br>✓ Cost-efficiency<br>✓ Transfer learning and fine-tuning<br>✓ Domain-specific adaptation<br>✓ Accelerates research<br>✓ Interactive learning tool<br>✓ Zero-shot and few-shot learning<br>✓ Creative writing aid<br>✓ Personalization<br>✓ Versatility<br>✓ Multilingual capabilities<br>✓ Time-saving | ✓ Continuous learning<br>✓ Language translation<br>✓ Code assistance<br>✓ Enhanced learning Experiences<br>✓ Natural language understanding<br>✓ Creative collaboration<br>✓ Mental health support<br>✓ Multilingual conversational abilities<br>✓ Context retention<br>✓ Accessibility<br>✓ Massive scale data processing<br>✓ Human-like interactions<br>✓ Adaptability to domains | ✓ Significant computational costs<br>✓ Misinformation propagation<br>✓ Ambiguity in responses<br>✓ Biased outputs<br>✓ Ethical concerns<br>✓ No real-time data<br>✓ Overfitting issues<br>✓ Lack of emotional intelligence<br>✓ Over-reliance on training data<br>✓ Over-reliance by users<br>✓ Vulnerability to prompt injection attacks<br>✓ Dependency on prompt phrasing | ✓ Limited context understanding<br>✓ Difficulty with complex queries<br>✓ Lack of real understanding<br>✓ Data privacy risks<br>✓ Security vulnerabilities<br>✓ Privacy risks<br>✓ Dependence on human moderation<br>✓ Limited contextual understanding<br>✓ Inability to reason causally<br>✓ Generating misinformation<br>✓ Dependency on training data |

**Fig. 7**. Benefits and challenges of using ChatGPT

### *3.1.2.6 Summarization*

Automatic Text Summarization (ATS) is a significant topic due to the vast volume of textual data that is rapidly growing nowadays on the internet and other sources. Three ATS approaches are available: abstractive, extractive, and hybrid [193]. To create a summary, the extractive technique chooses and incorporates the most significant sentences from the source text. In contrast, the abstractive approach transforms the input content into an intermediate representation before generating a summary using unique terms. The hybrid approach incorporates both the extractive and abstractive processes. L. Chen and Nguyen [194] proposed an ATS technique for summarizing sign documents based on a Reinforcement Learning (RL) technique and an encoder-extractor network architecture's RNN model. The key features are selected using a sentence-level encoding approach, after which summary sentences are obtained. However, the method has a chance of suffering from large variances since it uses an approximation in the RL method training objective function.

Clustering, unsupervised NNs, and topic modeling were used to create an Arabic summarization approach in this study [195]. The researchers also presented ensemble learning models and neural network models to combine the information produced by the topic space. Specifically, the ELM-AE method and k-mean approach were used to cluster documents on a large sample of Arabic documents. The LDA technique was implemented to determine the topic space associated with each cluster; then, the discovered subject space was used to generate a numerical document representation. For learning unsupervised features from texts, this format was utilized as the input for various NN and ensemble methods. For the variation of the information in the final summary, the learned characteristics were utilized to rank phrases by following a graph-based model. Key phrases were picked by applying the redundancy removal components. A few more unsupervised NN methods, for instance, stacked-auto-encoder, could be included in future studies to improve the robustness of the suggested technique.

### 3..1.2.7 Advanced text mining and classification

A methodology named heterogeneous latent topic discovery (HLTD) was proposed by Li et al. [196] that combines word embeddings with topic modeling to make use of both local word sequential relationships and global document topic information. By allocating word distributions according to document topics or word embeddings, HLTD improves the framework's all-encompassing perspective of the text. For parameter and hidden variable inference inside HLTD, a privacy-preserving Monte Carlo expectation-maximization technique is suggested to guarantee data privacy. By integrating differential privacy with conventional Markov Chain Monte Carlo (MCMC) methods, this approach introduced two new sampling algorithms that maintain model performance while safeguarding the privacy of training data. These algorithms can be used in various contexts and incorporated with others that deal with word embedding.

A model agnostic approach, variational continuous label distribution learning (VCLDL), was developed by Zhao et al. [197] to tackle the class imbalance issue in multi-label text classification (MLTC). To mine the data concealed in the observable logical labels, VCLDL creates a suitable link between the feature space and the label space. In particular, VCLDL treats the label distribution as a continuous density function in latent space and develops a versatile variational method to jointly use the feature space and estimate the label density function. When compared to the currently available MLTC models, the experimental results on various benchmark datasets revealed that VCLDL could significantly improve performance. Incorporating VCLDL into MLTC models allows them to focus on the distribution of the entire set of labels instead of individual labels with high response values, effectively resolving the class imbalance issue.

### 3.1.3 Computer vision

To a broader extent, computer vision is an interdisciplinary branch of computer science study that examines how computers can quickly recognize digital images and movies. The predominant processes include extracting, analyzing, and comprehending relevant information autonomously about a specific picture or a series of images [198]. Computer vision implies the algorithmic and theoretical roots for spontaneous ocular comprehension. The growth of computer vision is reliant on the computer technology system, focusing on image quality enhancement or image classification [199]. Researchers often use the terms interchangeably in the basic approaches because of the overlapping with image processing. However, the core aim of computer vision is to develop models, data extraction, and information from images. On the other hand, image rectification deals with applying computational alterations to pictures, such as sharpening and contrast, among other things [200]. Deep learning algorithms recently tackled major computer vision tasks like face recognition, object detection, human pose estimation, activity recognition, and fruit defect detection [201].

### 3.1.3.1 Object detection

Object detection has attracted massive attention as a vital function in computer vision, in which CNN has made substantial headway [202]. Object detection is, in essence, the first step in several computer vision applications, including logo detection, face recognition, pedestrian identification, and video analysis. One-stage detectors (YOLO and its variants) and two-stage detectors (region-based CNN (R-CNN)) are the two primary families of object detection deep learning frameworks [203]. One-stage detectors do not need the cascaded zone classification step to make qualitative predictions of objects on every position of feature maps. Two-stage detectors, on the other hand, have a proposal generator that makes a handful of suggestions before feature extraction. When dealing with complicated models, deep neural architectures are superior to simple ones [204]. Using the weight-sharing principle, CNN is a feed-forward neural network. CNNs are not precise for minor data, but produce high accuracy for vast image datasets [205]. However, to execute computer vision tasks, CNNs require large labeled datasets. It is a fusion of multiple functions and integration that shows how they intersect. The CNN layered structure for object detection is shown in **Fig. 8**.

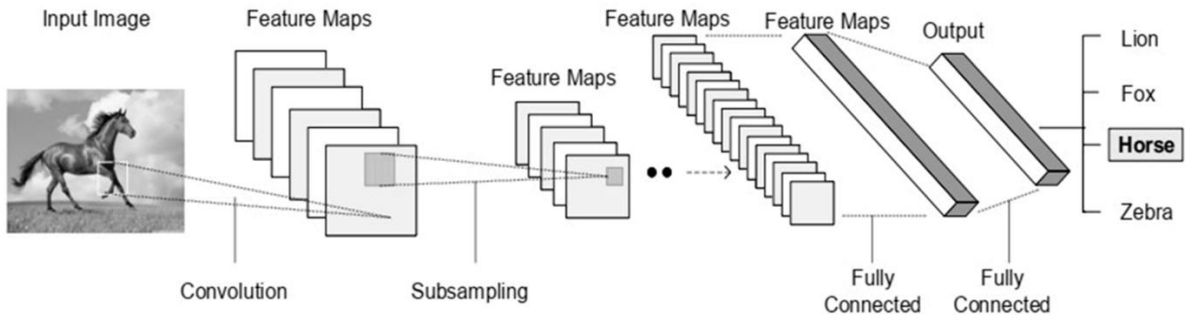

**Fig. 8.** Layered CNN structure for object detection (reprinted with the permission of Elsevier from [206])

For object detection, deep R-CNNs have been widely deployed. Ouyang et al. [207] suggested a deformable deep CNN for generic object detection. This constrained pooling layer in the recommended deep design simulates the deformation of object portions with geometric compensation and limitation. The researchers introduced diverse models by changing the training process, net topologies, and altering some vital components in the detection pipeline. This strategy exceptionally optimized the model's averaging efficacy. The proposed technique elevated the mean averaged precision by RCNN from 31 to 50.3, exceeding GoogLeNet, the winner of the ILSVRC2014, by 6.1%. The extensive experimental evaluation also provides a detailed component analysis. Doulamis and Doulami [208] explored a deep model's representation capacities in a semi-supervised scenario. The method overcame deep learning's crucial drawbacks by utilizing unsupervised data to construct the network and then fine-tuning the data using a gradient descent optimization process. Unsupervised learning appreciates more abstract representations to reduce the input data, thus optimizing the model's stability, accuracy, and reliability. In addition, an adaptive approach allows modifying the model dynamically to the digital photogrammetry conditions.

### 3.1.3.2 Face recognition

Given the enormous diversity of face images present in the real world, face recognition is amongst the most complex biometrics attributes to use in open circumstances [209]. Face recognition structures are frequently comprised of several components, such as an image, face detection, face representation, face matching, and face alignment. Researchers have focused on specialized approaches for each face variation owing to the feature's sturdy complexity in unconstrained environments. Many new face recognition architectures have been presented in tandem with deep learning advances, and some even come near to human performance. Extensive research with acceptable outcomes has been conducted on the issues of illumination, position, and expression [210]. However, when dealing with noisy photos, the precision of most methods deteriorates dramatically. Humans can often have a hard time distinguishing an identity from a seriously noisy face.

Zadeh et al. [211] designed a local model with convolutional specialists' constraints for facial landmark detection by adding a set of appearance prototypes of diverse poses and expressions. A pivotal part of the CE-CLMs is the Convolutional Experts Network (CEN), a new local sensor that blends neural structures with expertise in an end-to-end framework. The findings suggest that launching from iteratively-stated CEN network weights while exerting Menpo trained data alone does not yield competent outcomes. However, applying plain datasets of the Curriculum Learning paradigm [212] and switching to Menpo training data gave satisfactory results. Deng et al. [213] suggested UV-generative adversarial networks, a painstakingly built system that unifies global and local adversarial deep CNNs to develop an identity-preserving facial UV closure framework for pose-invariant visual recognition. Combining posture augmentation during training and pose discrepancy depletion throughout testing, the protocol attained a state-of-the-art affirmation accuracy of 94.05%.

To adversarially train the identity-distilled attributes, Y. Liu et al. [214] built an autoencoder system with minimal direction via facial identities. The model was shown to be capable of generating identity-distilled features and extracting concealed identity-dispelled features that can preserve complementary knowledge, such as backdrop clutters and intra-personal variances. The primary limitation of deep learning approaches is that they must be learned on extremely big datasets with enough variety to generalize to unknown samples. Recently, numerous large-scale face datasets, including images of faces in the wild, have been made available to train CNN models [215, 216]. The studies demonstrate that neural networks may get trained as classifiers and can reduce dimensionality.

An innovative model for detecting and recognizing faces on drones was proposed by Rostami et al. [217] to enhance face recognition performance in situations where query pictures are captured from long distances or high altitudes, resulting in limited facial information of the individuals. The authors also aimed to improve the level of performance by using a deep neural network to carry out these tasks. The proposed method was deconstructed into four phases: pre-processing, detecting faces, extracting features, and recognizing faces, as shown in **Fig. 9**. A deep CNN model was employed to localize faces in face detection, which was then used in face recognition. The process of feature extraction involved the extraction of deep features from the generated face images for use in the subsequent step. The process of face classification was conducted using CNN, which entails identifying a face from a series of facial images. Experimental testing on the DroneFace dataset compared the suggested framework to the state-of-the-art models, showing that the suggested approach can achieve competitive detection and recognition accuracies. Even when unmanned aerial vehicles (UAVs) take pictures with a wide angle of depression, the suggested method can still identify faces on them within certain angle limits. To overcome this limitation, the developed model can use the grid concept. It primarily models geometric facial components, including the mouth, eyes, and cheeks.

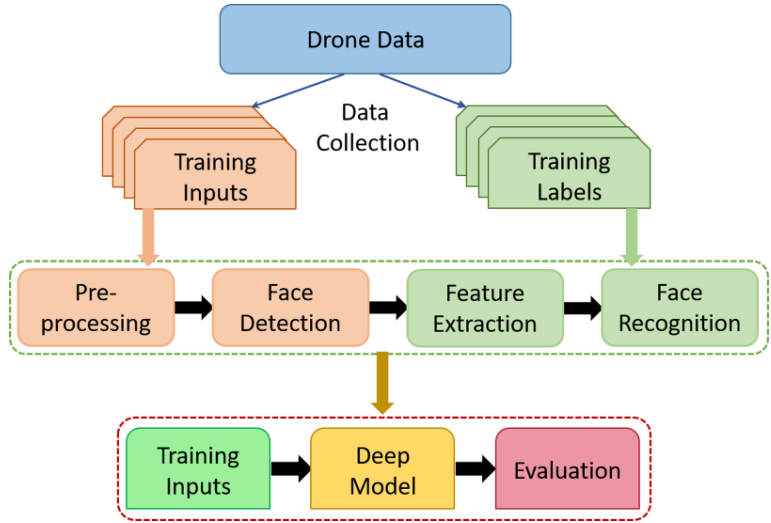

**Fig. 9.** Flowchart of the proposed framework (modified from [217]

### 3.1.3.3 Action and activity recognition

Human activity recognition (HAR) plays a vital part in today's society because of its potential to assimilate extensive advanced information from raw sensor data about human actions and activities [54]. To identify sensor acquired data, early studies primarily applied naive bias, decision trees, SVM, and other classic machine learning methodologies [218, 219]. Researchers have recently transitioned from traditional handcrafting to deep learning methods for HAR, especially with the advent and efficient application of deep learning algorithms. Handcrafted model representation cannot tackle complicated cases due to their limitations. However, given the success of DL models in speech recognition, NLP, image classification, and other domains, transferring them to the HAR field is a new research concept for pattern identification [220, 221].

Ordóñez et al. [222] reported a classifier that can classify 5 movements and 27 hand gestures using CNN and LSTM. The methodology surpasses deep non-recurrent networks by an average of 4% and outperformed some previous findings by 9%. Results suggest that this model can be used with homogenous sensor modalities and can fuse multimodal sensors for better execution. To reliably classify subjects or activities, Lin et al. [221] introduced a novel iterative CNN technique using autocorrelation pre-processing rather than the conventional micro-Doppler image pre-processing. An iterative DL framework was used in the proposed method to define and extract features automatically. In addition to outperforming feature-based methods employing micro-Doppler pictures, the proposed iterative CNNs, followed by random forests, also outperform classification methods employing various types of supervised classifiers.

Even though the preceding models may distinguish human activities in general, the entire network structure is quite complex. Furthermore, the large number of parameters has a massive computational cost in these models, which is thus challenging to employ in situations where high real-time performance is required. Agarwal and Alam [223] implemented a lightweight deep learning paradigm for HAR on a Raspberry Pi3. The overall accuracy of this model on the WISDM dataset was 95.78% using a simplistic RNN incorporating the LSTM technique. Despite its excellent accuracy and simplicity, the conceptual framework was evaluated on a single dataset with only six activities, which does not guarantee that it is generalizable. To address the drawbacks, Xia et al. [224] introduced a unique deep LSTM-CNN for HAR that can extricate activity features and categorize them automatically using only a few factors. The network was tested with three of the most often utilized publicly available datasets. The analysis showed that the network not only has fewer parameters with great precision but also has fast convergence speed with sufficient generalization aptitude. In many computer vision applications, deep learning has outperformed older methodologies due to the capacity of DL algorithms to learn characteristics from raw data, depleting the need for constructed feature detectors and descriptors.

### 3.1.3.4 Human pose estimation

The goal of the human pose estimation (HPE) problem, which has existed for centuries, is to determine how well sensor inputs can be used to estimate human posture. The HPE system attempts to determine a human's body position from still images or moving video. It is an important research field because it involves various applications, such as action detection, activity recognition, movies, video surveillance, human tracking, sports motion, virtual reality, medical assistance, human-computer communication, and self-driving analysis. HPE is challenging because of the wide variety of body types, self-obscuring stances, and complex environments that might occur from the great degree of freedom of interconnected joints and limbs [225]. Toshev and Szegedy [226] initially tried to instruct an AlexNet-like DNN to determine joint points from complete pictures without utilizing any models or part detectors. A multi-stage refining regressor cascade architecture improved the performance and modified the cropped images from the previous step. Similarly, Pfister et al. [227] used a string of concatenated frames to implement an AlexNet-like network to forecast the human stance in films. The use of joints alone without the context is not robust; however, turning them into numerical-joint positions from heatmap supervision monitoring can benefit both approaches.

Yang et al. [228] developed a Pyramid Residual Module (PRM) to increase Deep CNN (DCNN) stability across levels to swap the Hourglass network's residual module. This method exhibited considerably enhanced performance compared to the earlier state-of-the-art procedures. Papandreou et al. [229] developed a box-free, multi-tasking ResNet-based network for pose evaluation and occurrence classification. In real-time, the ResNet-based network can forecast joint heatmaps of all critical spots of all people, as well as their relative displacements. Then, the most confident detection is grouped, adopting a decoding strategy focused on a tree-structured kinematic graph. Furthermore, the network acquired a greater average precision and above 5% absolute advancement over the former top-performing approach on a similar dataset. Other studies [230–233] mainly focused on moderately constrained directions, such as 3D HPE, RGB-D-based action identification, body parts-based HPE, and monocular-based HPE model-based HPE.

## 3.2 Environmental and social

### 3.2.1 Ecology

The endless data stream offers ecologists a new challenge: finding or developing the analytical models required for data extraction from the massive amounts of video feeds and streaming photos [234]. On the other hand, obtaining usable footage in maritime zones to attain acceptable computing performance provides a unique set of obstacles compared to terrestrial situations. Variable water purity, obstruction due to schooling fish, complicated background formations, and diminished light with increased depth are some environmental complications that might interfere with clear footage in aquatic ecosystems [235]. Though certain elements may affect the image and video quality, deep learning techniques have evinced effectiveness in a variety of marine implementations. Deep learning algorithms are currently pirouetted around marine environments to automate the classification of certain species. Not surprisingly, the most common use of deep learning is identifying species from recordings of sounds, images, or videos. These investigations already cover a wide variety of organisms, from bacteria and protozoa to plants, insects, and vertebrates, both living and extinct, and from microscopic to planetary levels [236–246].

Deep learning's capacity to comprehend features from data is what deems it so powerful. Unsupervised algorithms have no specified output and are frequently used as a systematic appliance to find patterns in data, minimize dimensions, and classify groupings [234]. Ditria et al. [247] emulated the speed and precision of deep learning approaches opposite to human counterparts for verifying fish abundance in underwater video clips and photos to test its adequacy and relevance. They designed three prototypes using Mask R-CNN, an object detection framework to locate the aspired species called luderick (*Girella tricuspidata*). In single image test datasets, the machine topped marine specialists by 7.1% and citizen scientists by 13.4% and outperformed in video datasets by 1.5% and 7.8%, respectively. These results confirm that deep learning is a better tool to evaluate abundance with stable results and is portable across survey locations than humans.

Another vast potential of deep learning is to detect disease symptoms similar to existing applications in the medical sector. For example, CNNs can identify tree defoliation and crop illnesses. Rather than accounting for each feature independently through explicit modeling, Kälin et al. [248] built a joint distribution of techniques, applying 5-fold cross-validation for the preliminary investigation to eradicate any train-test split bias. Defoliation frequencies were distributed evenly among all 5 folds. On one of the datasets, this technique performed only 0.9% less than a group of human experts. This protocol can spot malnutrition, scars, and the presence of apparent diseases in wild plants and animals. Although species-specific models might forecast tree stress in a better way, the study does not have enough training data for each species to build a robust algorithm. In consequence, complex situations, including abundantly defoliated trees or a green canopy with trivial defoliation, can produce errors.

Traditional methods for classifying coastal pollution suffer from inefficiency and poor accuracy. A DCNN based on deep learning was proposed by Sathish et al. [249] to address these problems. The image's undesired noise was removed using anisotropic diffusion. To improve the picture's contrast, global histogram equalization (GHE) was used. After the image underwent pre-processing, it was analyzed and classified using a deep learning-based DCNN to identify and assess coastal pollution. The DCNN's weights were fine-tuned for optimal performance in classification by employing a reptile search algorithm. The suggested method was found to be 97.2% accurate, 96% precise, 2.8% error, 95.4% recall, 96.5% f1-score, and 96% specific. When tested against other methods, the developed approach performed better. In order to maintain environmental health and ecological safety, this model aids in the detection of coastal pollution.

Humans can abstract widely between topics and make accurate judgments with minimal information. However, deep learning algorithms have limited abstraction and reasoning capabilities. Only a few researchers have looked into combining different plant organs or perspectives to improve accuracy [250, 251]. Owing to the high effort required to collect and label the datasets, completely automated species identification remains a long way off. The excellence of an automatic detection model is determined not only by the quantity but also by the caliber of the given training data [252]. On the contrary, most of the reviewed research indicates a deficit in the available qualifying data.

### 3.2.2 Agriculture

The successful application of deep learning has spread to many fields, including agriculture. Some specific problems in this field that can be explored with deep learning methods include land cover categorization, plant classification, fruit counting, and crop type classifications. In a survey, Kamilaris and Prenafeta-Boldú [253] found that deep learning outperforms frequently-used image processing approaches in terms of accuracy in the agricultural domain. Many methods based on deep learning have been developed recently for identifying leaf stress in plants. Ramcharan et al. [254] offered a cassava vegetable-based transfer learning approach. An Inception V3 model was used on a sample of 15,000 photos, and the approach outperformed more common models based on machine learning, such as K-Nearest Neighbor (KNN) and SVM, with an accuracy of 93%. For certain diseases (CBSD, BLS, GMD), using the leaflet rather than the whole leaf increased diagnostic accuracy. However, whole-leaf photos enhanced accuracy for diseases like CMD and RMD.

Plant diseases are responsible for a significant portion of the agricultural produce loss. Reduced agricultural production might worsen the global food crisis and significantly impact national economies. Therefore, preventing the spread of plant diseases through early diagnosis is crucial to maintaining global health. This motivation drives scientists to develop automated systems to diagnose plant diseases early. To improve the accuracy of plant leaf disease classification, Ali et al. [255] utilized an ensemble of several deep learning models, including DenseNet201, inceptionresnetV2, efficientNetB0, and efficientNetB3. An innovative method for processing images was proposed by the authors to make deep-learning models more effective. PlantVillage, the largest dataset for plant disease, was used to train and test five different DL models. To find the most accurate DL ensemble, ten randomly selected DL model ensembles were selected to test and compare them. On the dataset, the suggested ensemble model achieved an accuracy of 99.89%. This high accuracy demonstrates the system's generalizability to new datasets or real-world situations.

Instead of evaluating the entire leaf, Arnal Barbedo [256] employed a CNN to categorize specific diseases and spots on plant leaves. This revealed a number of illnesses that all damage the same leaf. In addition, the author used deep learning to recognize specific lesions and spots in 14 plant species. The models in this study were trained using a pre-trained GoogLeNet CNN. The study could not determine the number of pictures of disease symptoms needed for the neural network to learn their characteristics. In all situations, a few hundred photos appeared to be sufficient to produce credible findings, but this quantity must be approached with caution.

Lin et al. [257] applied an RGB sensor along with a UNet-CNN to classify and identify cucumber powdery mildew-affected leaves. Because the loss value of the affected pixels in this study was lower than the non-affected pixels, the experiment presented a loss function (Binary-Cross-Entropy) to tenfold the loss values. The semantic segmentation CNN model segmented the sick powdery mildew on cucumber leaf pictures with a mean pixel accuracy of 96.08%. However, taking images in a controlled setting is necessary rather than in an open field. Furthermore, a lack of appropriate amount and diversity of the datasets, where symptoms produced by other conditions were not included, may hinder the effectiveness of DL methods.

DL approaches have recently been implemented in smart fish farms. A deep learning method was demonstrated to differentiate better variations in traits, classes, and the environment, allowing it to extract target fish attributes from a picture captured in an uncontrolled underwater environment [258]. On the well-known LifeCLEF-14 and LifeCLEF-15 fish datasets, CNN outperformed conventional techniques with classification accuracies of over 90% [259]. General deep structures in the experiment should be fine-tuned to increase the efficacy of identifying vital information in the feature space of interest, thus reducing the requirement for vast volumes of annotated data.

### 3.2.3 Financial fraud detection

Financial fraud is a type of thievery whereby a business entity or anyone illegally takes money or assets to profit from it. Since the advent of new technologies, there has been a massive rise in financial fraud involving all aspects of the business world. For instance, financial fraud victims in the United States lost an average of $1,090, accumulating over $3.2 billion in 2017 [260]. The selection procedure for practitioners is frequently known as safeguards of fraud analysis, assessment, or detection by information security specialists and associations. Some methods for detecting anomalies include decision trees, logistic regression, SVM, and others. However, these methods are constrained as supervised algorithms rely on labels to determine transaction validity [261].

Several studies [262–264] revealed that developing measures based on fraud estimations has been an infuriating procedure that is excessive and futile when tackling composite non-linear data, has inefficient memory utilization, lengthy calculation, and generates abundant false alarms. Financial fraud detection is still strenuous because of the constant changes in fraudulent behavior, the absence of a mechanism to track fraudulent transactions' information, certain limitations of the existing detection techniques, and highly skewed datasets [265]. Therefore, optimization of previous procedures and novel approaches are necessary to increase fraud detection rates [266, 267]. As a result, deep learning, a subset of modern artificial intelligence, is a promising technique recognized in recent years. DL also facilitates computational models with multiple processing layers to adapt data representation to different abstraction levels.

Credit card fraud is one of the most pervasive forms of financial fraud in which a hacker gains unauthorized access to a legitimate card without the cardholder's consensus. The simultaneous expansion of e-commerce and the internet has resulted in a significant increase in credit card use, resulting in an unusual increase in credit card theft in recent years. For instance, a report by the Boston Consulting Group showed that North American financial institutions lost $3 billion in 2017 due to credit card fraud. Despite industry attempts to combat economic fraud, the Nilson Report site revealed that global financial losses exceeded $24.71 billion in 2016 and $27.69 billion in 2017, attributed to credit card fraud [268]. These fraudulent occurrences may erode consumer confidence, destabilize economies, and elevate people's living costs. As a result, several studies have applied deep learning to detect credit card fraud.

Over 80 million credit card transactions labeled as fraudulent or legitimate were used to assess the efficacy of the deep learning process reported by Roy et al. [269]. This study assessed the performance of various DL algorithms based on their parameters, class imbalance, scalability, and sensitivity analysis. The researchers used LSTM, Gated Recurrent Unit, RNN, and ANN in a distributed cloud computing environment. The findings confirm that the LSTM technique achieved the best performance and reveal that each of the four topologies' execution expanded with the increased network size. However, the study failed to specify the conditions under which the model functions cease to ameliorate, model sensitivity, and limit the network size.

Gated architectures, such as LSTMs and GRUs, excel in sequential learning tasks, making them invaluable for processing large datasets with sequential or temporal dependencies [270]. They address challenges like vanishing and exploding gradients, common in traditional RNNs, ensuring stable and efficient training [222]. They are crucial in unraveling complex data by capturing long-term dependencies, adapting to variable-length sequences, and serving as effective feature extractors. Their ability to automatically learn and emphasize salient features within the data is instrumental for discerning patterns and making informed predictions. As versatile tools, gated architectures contribute significantly to analyzing complicated data structures, allowing for a more nuanced understanding of complex information.

To improve upon existing detection strategies and increase the detection accuracy of complex data sets, Alghofaili et al. [265] introduced an LSTM-based fraud detection method. The system was tested with a unique dataset of credit card frauds. Simultaneously, the outputs are analogized to an auto-encoder model and machine learning technologies in a pre-existing deep learning model to detect suspicious financial activities and alert the appropriate authorities, allowing them to take appropriate action. Although some machine learning techniques have demonstrated promising upshots, they do not detect new patterns or deal with large amounts of data to improve accuracy after reaching a saturation point. LSTM worked brilliantly in the trials, reaching 99.95% accuracy in less than a minute. The unique applicability of LSTM in prediction is ensured by prior experience and the correlation between prediction outputs and historical input. Many-state memory cells and gates are incorporated into the framework. Due to its vital role in ensuring that data is transmitted unmodified, the cell's state is at the core of the information transfer process [238].

Several studies used other deep learning frameworks productively for a similar cause. To illustrate, Pandey [271] introduced the H2O deep learning framework in credit card forgery detection. The pattern of the model enables multiple algorithms to aggregate as modules, and their outputs can be integrated to improve the final output accuracy. Concurrently, the efficiency of the algorithms rises as the dataset length grows while using this framework; thus, adding more algorithms with equivalent formats and datasets can elevate this model. Another type of financial fraud is tax fraud, the malicious act of falsifying a tax return document to reduce someone's taxation. Low fiscal revenues are weakening governmental investment [272]. When it comes to the property acquisition tax, Lee [273] implemented a reliable sampling technique for tax-paying citizens. The data from 2,228 returns were fed into autoencoder, a well-known unsupervised deep learning technique, to calculate the potential tax shortfalls for each return based on an estimate of the rebuilding faults. The sorted reconstruction ratings are compatible with practical context, implying that the defects can be exploited to identify suspicious payers for auditing in a cost-efficient strategy. Utilizing the recommended strategy in real-world tax administration can reinforce the self-assessment acquisition tax system.

López et al. [274] researched tax fraud detection in the context of individual income tax filings. Neural networks application enabled the taxpayers' segmentation and the possible assessment of a particular taxpayer attempting to avoid taxes. The neural network outcome categorizes whether a taxpayer is deceitful or not as well as reveals a taxpayer's proclivity for unethical activities. In other words, it classifies individuals based on their potential of convicting and also calculates the propensity of tax fraud per taxpayer. The chosen model excelled over other tax fraud detection algorithms with an accuracy of 84.3%. It would be interesting to see this concept applied to additional taxes in the future.

To improve the fraud detection of credit cards, Khalid et al. [275] proposed a new ensemble model combining several classifiers: SVM, KNN, Bagging, RF, and Boosting. The model was tested using a dataset that included credit card transaction records from European customers to ensure a realistic evaluation. With the computational capabilities of Google Colab, the proposed model's methodology included feature engineering, data pre-processing, model choice, and evaluation. This allowed for effective training and testing of the model. Compared to conventional ML techniques and individual classifiers, the suggested ensemble model has been demonstrated to be more effective in reducing difficulties related to fraud detection of credit cards. The ensemble achieved better results than the preexisting models in recall, accuracy, precision, and F1-score metrics.

### 3.2.4 Disaster management

In terms of disaster or natural calamity management, uncertainty, lack of resources in the affected areas, and dynamic environmental changes are the major characteristics of natural catastrophes. The inability to predict outcomes indicates that catastrophic effects on persons and property during several disasters cannot be foreseen with reasonable accuracy [276]. Big data is a technology paradigm that enables researchers to efficiently analyze enormous amounts of data accessible by current practices [277]. Making the most of big data is possible using various technological and scientific strategies and equipment. Recent advancements in big data and IoT technologies open up a wide range of opportunities for disaster management systems in order to obtain leading assistance, guidance, as well as better observations and ideas for precise and suitable decision-making.

In order to resolve catastrophic challenges, Anbarasan et al. [278] introduced concepts and strategies for identifying flood catastrophes based on IoT, big data, and CDNN. Big data derived from the flood catastrophe was first used as the input data source. Afterward, the Hadoop Distributed File System (HDFS) was used to decrease the high frequency of data. After excluding high-frequency data, the data were preprocessed via missing value approximation and a normalizing technique. A conjunction of attributes was utilized to construct the pattern centered on the pre-processed data. In the final step, the created parameters were sent into the CDNN classifier, which categorized them into two categories: the probabilities and infeasibility of a flood emerging. The performance of the suggested system was analyzed in terms of precision, accuracy, recall, F-score, specificity, and sensitivity compared to the existing systems, DNN, and artificial neural network (ANN). The comparison findings demonstrate that the CDNN method has a greater level of accuracy than the present approach, with 93.2% accuracy, 92.23% precision, 90.36 % recall, and 91.28% F-score with 500 data points utilizing CDNN. The results described above can be improved using DNN as well as ANN models. In conclusion, the detection system performed better than other current methods. In the future, the work presented here could be improved using IoT devices with even longer sensor ranges at lower costs and cutting-edge algorithms at each stage of the flood identification process.

Disasters caused by fire do have a negative impact on the environment, society, and economy. Early fire detection and an automated response are vital for disaster management schemes to minimize these damages. Fire detection early in disaster management systems while monitoring public spaces, woods, and nuclear power plants can prevent ecological, monetary, and social harm [279]. The movements of entities with a retardant appearance and varied atmospheric factors make early detection hard [280]. Subsequently, an improved accuracy algorithm that reduces the number of false alerts in the conditions above is required. In order to accomplish this, Khan et al. [281] investigated deep neural networks and developed a prominent architecture for early fire detection through surveillance for efficient disaster management systems. Another preferable criterion was to notify the disaster management system after a successful fire detection and include the representative frames. A system of flexible prioritization was developed for the monitoring system's camera nodes, taking into account the contents they could visualize. The experiment primarily examined two datasets. The initial dataset demonstrated additional progress, as accuracy increased from 93% to 94% and false positives decreased from 0.11% to 0.9%. Despite producing false negatives of around 21%, the approach maintained a more robust balance, enabling the method to achieve fire identification to achieve fire identification more accurately more accurately. The findings for the second dataset [282] were gathered using a different set of metrics, including recall, precision, and F-score, to properly conduct the performance. Results produced by using deep features and optimizing the fire prediction model are compared and contrasted, showing that the detection approach and model are feasible with a maximum score of 82% for precision, 98% for recall, and 89% for F-score. Despite evaluating the sustainability of the proposed approach utilizing noise intrusions, scaling, and rotations, each image comprised of fire and showed a high accuracy ranging from 89% to 99%. Considering the potential of such occurrences occurring under surveillance, the results indicate that the CNN-based model could detect fires at an early stage in a range of circumstances, even when images are blurry. The relevance of the framework for successful fire disaster management was supported by experimental results, which also confirm the high accuracy of the fire detection technique compared to cutting-edge methods.

Over the past few years, social media networks have contributed to managing natural catastrophes. In order to interpret better messages and extract relevant information from social media, text data mining techniques employing standard machine learning approaches have been constructed. These techniques are distinct and challenging to generalize for different categorizations. Therefore, considering the crisis management efforts related to hurricanes, Yu et al. [51] investigated the ability of a convolutional neural-based deep learning model to classify the trending catastrophic topics from Twitter. SVM and LR are two conventional machine learning techniques that were evaluated against the logistic regression of the CNN model. The outcomes of the experiment demonstrate that CNN models consistently outperformed SVM and LR models in terms of accuracy for both types of assessment scenarios. Moreover, the CNN classifier surpassed the classification methods of SVM (63-72%) and LR (44-60%), achieving an accuracy of up to 81% among all datasets. In order to categorize tweets posted during a later event, the evaluation was carried out using CNN based on Twitter data from prior events. Results of the experiment demonstrated that while SVM and LR's accuracy dramatically decreased, CNN maintained a steady performance. This suggests that the CNN model could be trained in advance using Twitter data from previous events to categorize new occurrences for situational awareness. However, CNN took longer to learn than SVM and LR because there were more parameters to consider, which makes it difficult to employ CNN for web-based learning.

When a disaster occurs and geotagging data associated with images shared on social media is typically unavailable, the framework proposed by Sathianarayanan et al. [283] can extract geographic locations from images using phone numbers. It consists of three main modules: phone number identification, image collection, and location extraction through Google Maps API. Using the Street View House Numbers (SVHN) and manually interpreted multi-digit phone numbers, the authors trained a CNN-based detection model—specifically, the RetinaNet object recognition algorithm—to identify and extract these numbers from images. The phone number identification model was able to achieve an Average Precision (AP) of all digits greater than 79% (0.79) and a respectable mean AP of 82% (0.8248) using an IoU (Intersection over Union) threshold of 0.5. Utilizing phone numbers retrieved from object detectors, the Google Maps API can render location data with reduced distance distortion. All kinds of disasters, including earthquakes, floods, and wildfires, can benefit from the proposed method's enhanced disaster assessment, situational awareness, and management capabilities.

### 3.3 Industrial and engineering

#### 3.3.1 Fluid dynamics

Fluid dynamics is an area of applied science dealing with the flow of liquids and gases. Three conservation laws define fluid dynamics: mass conservation, linear momentum conservation, and energy conservation. Computational fluid dynamics (CFD) is a set of numerical methods for providing a rough solution to fluid dynamics and thermal difficulties. CFD is not a science in itself but rather a means of applying numerical analysis principles to heat and mass transfer [284]. It analyzes airflow motion, contaminant transport, and heat transfer in enclosed places. CFD helps with wind flow and contamination dispersal analysis surrounding buildings in metropolitan surroundings but still confronts several difficulties concerning computing cost and accuracy.

The neoteric success of DNNs has been facilitated by a high abundance of computational power that takes advantage of the multi-layer architecture. Not long ago, DNNs gained notoriety in turbulence modeling or, comprehensively, in high-dimensional areas, complex dynamical systems [45]. Fonda et al. [285] scrutinized heat transport characteristics of turbulent Rayleigh–Bénard convection in horizontally expanded systems by using deep-learning frameworks. The researchers applied the deep CNN-trained method to measure the heat transfer's fraction and time variations. The slowly evolving turbulent superstructures received special attention because they are larger than the height of the convection layer. The strategy trains a deep CNN with a U-shaped configuration that has a contraction branch and an expansion branch, simplifying the complicated 3D superstructure in the midplane layer to a temporal planar network. As a result, data compression happens when the maximum Rayleigh number is more than five magnitudes. This shows deep learning's utility in parameterizing convection in global models of stellar and atmospheric convection. The U-specialized net's architecture requires fairly minimal training datasets that proved to be the most effective for ridge extraction, especially for noisier data at higher Rayleigh numbers. Nonetheless, the study did not provide information on the network's efficiency at lower Rayleigh numbers.

Another study conducted by Daw et al. [286] applied extensive fluid flow simulations to climatology and turbulence to forecast turbulent flow by identifying exceptionally nonlinear phenomena from spatiotemporal velocity fields. Specifically, they introduced adaptable spectral filters along with a specific U-net for assertion. The concept exhibits substantial declines in error prediction compared to state-of-the-art baselines. Most crucially, this procedure accurately previses over physical entities that satisfy favorable physical attributes, such as the conservation of mass, and also simulates the acute turbulent kinetic energy field and spectra for precise predictions. Another novel approach in this discipline is model recognition of reduced-order fluid dynamics process applying deep learning, which is a computational simulation-based method to create mathematical patterns of dynamic physical systems.

Reduced-order modeling (ROM) is one of the effective system identification strategies for reducing the complex and excessive dimensional size of discrete dynamical systems. Air pollution modeling, nonlinear large-scale systems, optimal control, ocean modeling, form optimization, neutron difficulties, sensor placement optimization, porous media problems, aerospace, multiscale fracture, and shallow water are some examples of its usage [47, 287–294]. Subsequently, Wang et al. [47] developed a ROM initially by combining a deep learning algorithm with suitable orthogonal decomposition methodologies. The results revealed that the DL ROM (DLROM) can capture sophisticated fluid dynamics with a CPU cost of less than 1/1000. While DLROM offers better prediction potential than earlier ROMs [295], deep learning frequently demands massive training data that significantly surpasses the network's parameters. When the probability distributions of the novel input data deviate from those of the training data, the resulting models are typically acceptable for interpolation but may not be adequate for extrapolation. Nevertheless, the effects of expanding this method to variable parametric obstacles, such as the real-time response to natural disasters, are unknown.

In order to achieve super-resolution of flow-field data in both time and space from a small set of noisy measurements in the absence of high-resolution historical data, physics-informed neural networks (PINNs) were employed in the study of Eivazi et al. [296]. Finding a physically consistent forecast at any location in the solution domain—a continuous solution to the problem—was the primary goal of the study. Three standard examples show that PINNs can be used to super-resolve time and space flow-field data: Burgers' equation, two-dimensional vortex shedding inside a circular cylinder, and minimum turbulent flow through a channel. A real-world experimental dataset, including hot-wire-anemometry measurements, demonstrated that PINNs' capabilities improved resolution and reduced noise. The results also demonstrated that PINNs are capable of handling data augmentation adequately.

### 3.3.2 Civil engineering

Deep learning is currently exerting a substantial influence in the domain of civil engineering, specifically in the realm of structural health monitoring (SHM) [297]. SHM systems have been extensively used on a variety of civil infrastructures to detect abnormalities or damage to structures and to track their overall structural health. These systems monitor environmental conditions, structural loadings, and responses over an extended period of time [48]. Many sensors are integrated into an SHM system to collect massive monitoring data that can accurately represent the operational state of the target structure. Deep neural networks are capable of real-time assessment of the structural integrity and detection of anomalies in sensor data originating from infrastructure such as dams and bridges. By taking this proactive stance, engineers are able to detect potential problems prior to their escalation, which ultimately results in improved safety and more effective maintenance strategies.

Early detection of damage constitutes the first stage of monitoring structural health. A new damage index was defined by Entezami et al. [298] using some empirical measures and the notion of the lowest distance value. This method was proposed within the theory of empirical ML and operates in an unsupervised learning method. This non-parametric approach successfully distinguishes between damaged and undamaged states with a high degree of damage detectability and negligible false positive/negative errors. The comparative studies also found that this method outperformed the anomaly detection approaches. When compared to the anomaly detection methods examined in the studies, this method also performs better.

On the basis of unsupervised anomaly detection, a probabilistic ML approach for civil structure health monitoring was proposed by Sarmadi and Yuen [299]. Among the several advantages of the suggested approach are its ability to estimate a trustworthy threshold, prepare discriminative novelty ratings for damage detection, handle the challenging problem of environmental and/or functional variability, and integrate the decision-making and threshold-estimating processes into a single structure. The results revealed that the method is a dependable and impactful tool for monitoring the health of civil structures in different environmental and operating circumstances. However, data quality, the interpretability of complicated models, and the necessity to carefully assess model transferability across various time scales are potential obstacles to structural health monitoring.

When evaluating the time and money saved by utilizing high-performance concrete (HPC), it is essential to accurately predict its compressive strength. To reliably forecast the compressive strength of the HPC, Islam et al. [300] proposed four DL models, namely BiLSTM, GRU, CNN, and LSTM. The model was developed by utilizing a substantial database that contained information about fly ash, cement, coarse aggregate, age, water, sand, blast furnace slag (input variable), and compressive strength (output variable). The study employed a training dataset comprising 80% of the total, with a testing dataset comprising the remaining 20%. During training, DL models demonstrated an R-squared value of approximately 0.960; during testing, it was nearly 0.940. However, the GRU model outperforms other models with an R-squared value of 0.990 during training and nearly 0.961 during testing, indicating high accuracy in both operations. Predicting the compressive strength of HPC using this method can be useful for building environmentally friendly infrastructures without resorting to expensive and time-consuming experiments.

### 3.3.2 Transportation prediction

Advanced transportation operations and services are enabled by traffic prediction, which is a critical component of the Intelligent Transportation System (ITS) that is designed to address the growing traffic congestion issues. The approach used for forecasting traffic has substantially transformed in the past few decades, transitioning from basic statistical models to the complicated integration of various DL models. Previously, analytical or statistical methods were used to solve problems in this domain. These enhancements have aided traffic management and planning, raised transit road safety and security, reduced maintenance costs, improved public transportation and ride-sharing firm performance, and propelled driverless car development to a new level.

Zheng et al. [301] presented a DL model that automatically extracts fundamental properties of traffic flow data using hybrid and multiple-layer architectures. They designed an attention-based ConvLSTM module that extracts spatial and short-term temporal features using a CNN and LSTM network. By automatically employing various weights to flow sequences at different periods, the attention method was correctly constructed to discern the relevance of the sequences at several intervals. In order to investigate long-term temporal features, the researchers proposed a bidirectional-LSTM (Bi-LSTM) architecture that extracts regular (daily and weekly) periodical features in order to grab the variance trend of traffic flow in prior and posterior orientation. However, only relatively small and simple road networks were considered in the study, which are more complicated and large scale. As a result, CNN and Bi-LSTM methods might be unable to completely utilize traffic flow's complex and dynamic properties. A deep learning-based traffic flow prediction system is demonstrated in **Fig. 10**. Input data from traffic sensors and GPS devices are processed by a CNN-LSTM model. The model analyzes sequential data to predict congestion levels, which are visually represented on a map with color-coded regions indicating traffic density. The neural network architecture is outlined, showing data progression from input to output prediction.

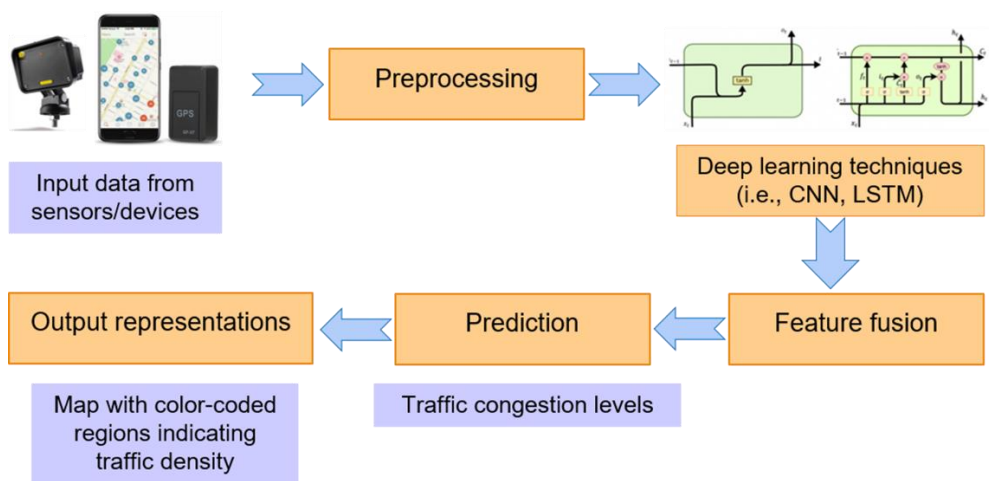

**Fig. 10**. Deep learning-based traffic flow prediction system

A multistep prediction model built on attention mechanism-based convolutional neural network (CNN)-LSTM was proposed by Vijayalakshmi et al. [302]. To increase model accuracy, the suggested technique employs the spatial and time-based features of data, which are retrieved by CNN and LSTM. This method facilitates the detection of short-term traffic characteristics (e.g., speed), which is essential for calculating future flow values. Weather and other conditions like accidents and road closure data could be taken into account for the reliability of the suggested technique. Abdollahi et al. [303] proposed a method for forecasting travel time using a multi-step process, which begins with removing both temporal and spatial outliers. To increase prediction accuracy, reduce overfitting risks, and achieve a more robust learner, the researchers employed a deep stacked autoencoder to present lower dimensions' features. While the proposed technique could capture traffic dynamics in general, it was not successful when heavy snow was present or another uncommon occurrence that had a substantial influence on travel times prevailed. To overcome this problem, several deep architectures and representation learning algorithms could be implemented, followed by a performance comparison to find an effective method.

The capacity to aggregate massive amounts of data from multiple sources for identifying events is a key feature of ITS. To address this problem, Reddy et al. [304] introduced a distributed fog-cloud ensemble that is IoT-integrated and context-aware. This ensemble intelligently handles context instances across fog nodes, assuring that the ITS has access to them. Employing a hybrid CNN system improved prediction accuracy, in which each vehicle only remembers local information, but nearby fog nodes can access global events through federated learning. The hybrid CNN model outperformed the LeNet by about 3% when tested with identical RGB input images; this results in an accuracy of over 95% in the experiments. The adoption of lightweight tracking schemes is required for ITS event detection due to extremely high latency requirements, which severely compromises the efficiency of ITS and makes prediction accuracy a challenge.

### 3.4 Biomedical

#### 3.4.1 Biomedicine

Many biological and medical data, including information about medical imaging, biological sequences, and protein structures, has been accumulated in recent decades due to the advancements in high-throughput technology. Consequently, deep learning has been extensively applied to biomedical data in biomedicine. For instance, CNNs are widely employed in biomedical image processing due to their extraordinary capacity to assess spatial features. CNNs have a lot of potential in omics analysis [57] and the study of biological signals [305], even though sequencing data employing CNNs is not very common. RNN-based architectures, on the other hand, are designed for sequential data and are more frequently utilized for transcriptome analysis [306] and dynamic biomedical signals [307]. Hence, the focus on deep learning is increasing in the field of biomedical information, and each paradigm may soon find new implementations.

#### 3.4.1.1 Prediction of protein structure

Protein structural analysis now relies heavily on cryo-electron microscopy (cryo-EM) [308], making atomic resolution possible. However, on all but the purest density maps with a resolution of less than 2.5 angstroms, estimating the structural trace of a protein remains difficult [309]. A DL model proposed by Si et al. [310] predicted the alpha carbon atoms throughout the core structure of proteins using a series of cascaded convolutional neural networks (C-CNNs). Using a semantic image classifier, the cascaded CNN was trained on a large number of generated concentration maps. With only a suggested threshold value needed per protein concentration map, this procedure was automatic and significant. To create the primary core trace with alpha carbon placements, a customized tabu-search path walking algorithm was applied. The alpha-helix secondary structural elements were further enhanced by a helix-refinement technique. Finally, to create full protein structures, a unique quality assessment-based creative approach was employed to successfully align protein sequences into alpha carbon traces, using 50 trial maps with resolutions between 2.6 and 4.4 angstroms. In terms of the proportion of connecting alpha carbon atoms, the proposed model produced more comprehensive core traces (88.9%), thus outperforming the Phoenix-based structure construction method with an accuracy of around 66.8%. By including additional protein structural details in the C-CNN or training the networks with experimental data, further study may enhance this research area.

In order to understand sequence patterns and folding structure, Park et al. [311] introduced deepMiRGene, which utilizes RNNs, particularly LSTM networks. The most significant contribution of this suggested approach is that it does not involve rigorous manual feature development. By utilizing end-to-end DL, this approach eliminates the need for extensive domain expertise and instead utilizes simple preprocessing. However, it is challenging to use an LSTM network right away due to the palindromic secondary structure of microRNA. To solve this problem, deepMiRGene employs a novel learning method in which the secondary structure from the input sequence is segmented into front and back streams. Additionally, deepMiRGene performed better than all other options in terms of sensitivity and specificity on testing datasets with higher sensitivity accuracies of 89%, 91%, and 88% in the method's three datasets. Even though there were significant disparities between the features of the various species, deepMiRGene also performed best when using cross-species data, thus demonstrating the potential for identifying inherent traits.

AlphaFold has revolutionized the field of structural biology by accurately predicting protein structures, a breakthrough that has propelled biomedical research into new territories. However, beyond AlphaFold, significant cutting-edge advancements are shaping the biomedical landscape. One notable example is the integration of generative AI models, such as large language models, into drug discovery pipelines. These models aid in predicting molecular interactions, designing novel compounds, and optimizing drug candidates, significantly reducing the time and costs associated with traditional drug development. Additionally, CRISPR-based genome editing technologies continue to expand their scope, enabling precise modifications at the genomic level to address previously untreatable genetic disorders.

AlphaFold by DeepMind revolutionized protein structure prediction by accurately modeling three-dimensional structures from amino acid sequences. This breakthrough addresses the long-standing protein-folding problem using advanced DL techniques. However, other cutting-edge innovations like RoseTTAFold, graph neural networks (GNNs), and multimodal techniques also merit discussion. A complementary approach to AlphaFold, RoseTTAFold utilizes a tri-branch architecture to integrate sequence information and structural predictions effectively. It shows promise in identifying protein complexes and interactions, which remain challenging for AlphaFold [312]. GNNs are increasingly applied in drug discovery to model molecular interactions and predict binding affinities, enabling the identification of novel therapeutics [313]. By incorporating these advancements, the biomedical domain is poised to address critical challenges, from precision medicine to understanding complex biological systems.

Another area witnessing rapid advancements is the use of personalized medicine, powered by multi-omics data integration and artificial intelligence. By combining genomics, proteomics, and metabolomics data, researchers can now create highly individualized therapeutic strategies tailored to a patient's unique molecular profile. In parallel, advancements in synthetic biology are enabling the development of living therapeutics, such as engineered bacterial or T-cell therapies, which can be programmed to target specific disease pathways with unprecedented precision.

Furthermore, advancements in neurotechnology, such as brain-computer interfaces, are opening possibilities for treating neurological disorders and restoring motor functions in paralyzed patients. Combined, these breakthroughs demonstrate that while AlphaFold has laid the groundwork for transforming biomedicine, a broader ecosystem of innovative technologies is driving the field toward a future of more precise, efficient, and personalized healthcare solutions.

With the development of AlphaFold [138], DeepMind has made a major contribution to computational biology by tackling the problem of protein structure prediction from amino acid sequences. Using advanced neural networks, AlphaFold predicts 3D protein structures very accurately. To describe the spatial configurations of amino acids, it combines evolutionary data with attention mechanisms. With this method, AlphaFold can learn patterns from massive volumes of sequence and structural data to predict complicated protein structures, often with near-experimental accuracy. With its ability to shed light on the molecular mechanics of proteins, AlphaFold has great promise for advancing biotechnology, expediting drug development, and improving our comprehension of biological processes.

A significant enhancement in the modeling accuracy was demonstrated in 2021 by the AlphaFold2 method, which led to a milestone in predicting protein structure. The AlphaFold Database made available to the public the predicted structures of proteins in 21 different species after the AlphaFold2 software was released. Aderinwale et al. [314] developed a tool called 3D-AF-Surfer to assist in protein structural analysis and utilization of AlphaFold2 models. This tool enables real-time structure-based searching for the AlphaFold2 models. The structures in 3D-AF-Surfer were modeled using 3D Zernike descriptors (3DZDs), which are mathematical representations of 3D shapes that are invariant to rotation. In the study, a neural network was built to take 3DZDs of proteins as input and achieved more precise retrieval of proteins with the same fold compared to 3DZDs' direct comparison. The 3D-AF-Surfer tool was utilized to classify the structures of AlphaFold2 models, and the relationship between the confidence levels of these models and intrinsically disordered regions was examined. The 3D-AF-Surfer demonstrated high accuracy while enabling real-time structure search, facilitating interactive analysis of Alphafold2 models.

A variety of predictors have been developed for specific interaction types, along with a generalist method, but their deep-learning accuracy often falls short compared to physics-based methods. These methods are typically specialized and cannot predict complex biomolecular structures involving multiple entity types. AlphaFold 3 (AF3) addresses this by accurately predicting complexes with nearly all molecular types found in the Protein Data Bank (PDB)[213]. It outperforms specialized methods in most categories, particularly in protein structure and protein-protein interactions, due to significant advancements in its architecture and training, as illustrated in **Fig. 11**.

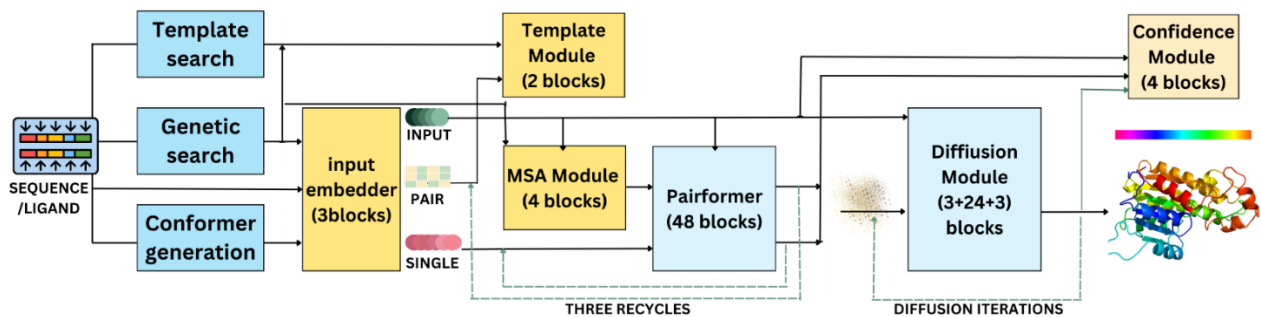

**Fig. 11**. AlphaFold3 architecture for the inference [315].

### 3.4.1.2 Genomic sequencing and gene expression analysis

Large-scale gene expression profiling has been frequently utilized in the characterization of cellular states in response to various illness circumstances, genetic mutations, and so on [316]. Creating a compilation consisting of thousands of samples of gene expression profiles is still quite unreasonable, even though the expense of entire gene expression profiles has been gradually declining. The computational strategy used by the LINCS program, however, currently relies on linear regression, which restricts its accuracy because it cannot detect complicated complex correlations among gene expressions.

Chen et al. [317] suggested using a DL technique called DGEX to predict the expression of key genes from the landmark gene profile expression. The model was trained using the microarray-based GEO dataset, which contains around 111,100 expression profiles, and its performance was evaluated against other approaches. With a 15.33% proportional improvement, deep learning greatly exceeded linear regression (LR) in terms of the average absolute error of mean through all of the genes. Deep learning outperformed LR in 99.97% of the key genes as per the gene-gene comparison investigation. A separate RNA-Seq-based GTEx dataset with around 20,000 expression profiles was also used to test the performance of the trained model.

### 3.4.1.3 Medical image classification and segmentation

Deep learning strategies have been utilized in studies segmenting cerebral tumors as a result of their achievement in general image analysis disciplines, including the classification of images [318] and semantic identification [319]. Conditional random fields (CRFs) and fully CNNs (FCNNs) in a single framework were proposed by Zhao et al. [320] to develop a unique deep learning-based identification approach for cerebral malignancy. The model was created to produce a malignancy segment that accurately identifies both appearance and concentration. Instead of employing CRFs as a step after FCNN post-processing, the authors utilized CRF-RNN to construct CRFs, making it simple to train both frameworks as a single deep network. The combined deep learning model was trained in three steps by employing segments and pixel patches. FCNNs were trained in the first step using image patches, while picture slices were used to train the next CRF-RNN in the second stage. In the third stage, the entire network was tuned using image slices. The unification of FCNNs and CRF-RNN, which acquired an accuracy of 88%, could increase segment resilience to model training factors, such as picture patch size and training picture patch count, as per the experimental data. Therefore, using 3D CRF as a step of post-processing could enhance malignancy identification performance.

Gliomas are the most prevalent and deadly type of cerebral tumor, with an extremely low survival probability in the maximum grade [321]. Planning the timeline of treatment is consequently essential for expanding the quality of life for cancer patients. The assessment of these tumors is frequently done using magnetic resonance imaging (MRI), but the volume of data generated by MRI makes it impossible to manually segment the images promptly. Accordingly, this restricts the use of profound quantitative assessments in clinical practice. The enormous spatial and functional heterogeneity among cerebral malignancies makes automatic segmentation a difficult task; hence, dependable and automated segmentation approaches are needed.

An automated segmentation technique based on CNN was suggested by Pereira et al. [322] to explore kernels for glioma segmentation in the images derived from MRI. As a result of the network's smaller number of weights, using small kernels enabled the design of more intricate architectures while also helping to prevent overfitting. Intensity normalization was also utilized as an essential step for pre-processing. While this is considered uncommon in CNN-based segmentation approaches, it proved successful when combined with data augmentation for segmenting brain tumors in images collected from MRI. By simultaneously receiving the top spot for the full, core, and expanding areas in the DSC measure (88%, 83%, and 77%), the proposed methodology demonstrated to be legitimate in the Cerebral Tumor Classification Challenge database.

### 3.4.2 Bioinformatics

The natural innate immune components known as antimicrobial peptides (AMPs) are commonly the focus of novel therapeutic developments due to the increment of antibiotic-resistant bacteria. Nowadays, wet-lab researchers frequently use machine learning techniques to find promising applicants. For instance, Veltri et al. [323] proposed a DL model to identify antibacterial activity by generating a neural network model that uses a core sequence composition and contains CCN and RNN layers. A comprehensive training and testing dataset incorporating the most recent antibacterial peptide information was utilized. In contrast to the current approaches, the proposed novel deep neural network (DNN) classifier performed better AMP identification. Through the implementation of DNN and RNN layers, the dependency on prior feature generation has decreased. Besides, a reduced-alphabet representation demonstrated that adequate AMP identification can be preserved using 9 different kinds of amino acids based on the embedded parameters. The DNN model performed the best in terms of MCC, auROC, and ACC and achieved a superior accuracy of 91.01% compared to other models, such as AntiBP2 (89.37%) and the CAMP Database RF model (87.57%). Thus, the suggested model eliminates the dependency on domain experts for feature creation by employing a deep network model that automatically extracts expert-free characteristics.

Nowadays, it is possible to measure DNA methylation at a single-cell level due to recent technological advancements. To facilitate genome-wide analysis, strategies to predict unknown methylation patterns are necessary because existing procedures are constrained by insufficient Chg. coverage. Angermueller et al. [324] developed DeepCpG, a computational method focusing on DNN to predict the methylation stages of a single cell and model the origins of DNA methylation dispersion. DeepCpG utilizes relationships between patterns of DNA sequences, stages of methylation, and nearby CpG sites both within and between cells. The extraction of useful characteristics and model training parts are not separated in the proposed method; rather, DeepCpG is built on a modular architecture and develops predicted sequences of DNA and patterns of methylation through data-based learning. Notably, DeepCpG performed better than RF-trained DNA and CpG characteristics. After training both the RF and DeepCpG models using only DNA sequence information, the highest relative improvements in the accuracy of 80% and 83% were achieved, respectively.

A novel deep learning technique named EPIVAN was introduced by Hong et al. [325] that allows lengthy EPIs to be predicted only from genomic sequences. Pre-trained DNA vectors were utilized to encode regulators and promoters in order to investigate the essential sequential properties. Subsequently, a one-dimensional convolutional model and deep neural units were used to identify both local and global characteristics to make a prediction regarding EPIs in different cell lines. The effectiveness of EPIVAN was compared to that of the neural network models, such as SPEID, EPIANN, and SIMCNN. Each model applied the identical test and training set to every cell line. On six cell lines, the proposed model's results and the four predictors were provided in terms of AUPR and AUROC. EPIVAN demonstrated the strongest AUROC of any model, with an accuracy ranging from 96.5% to 98.5%. According to test findings on six cell lines, EPIVAN outperformed the existing models, showed the capability of being utilized as a pre-trained model for further transfer learning, and demonstrated strong transfer ability.

Prediction of the functionality of enzymes is an important step in constructing new enzymes and diagnosing disorders connected to enzymes, which is a major task in bioinformatics [326]. Several studies have generally concentrated on predicting the mechanism of the monofunctional enzyme. However, the amount of multi-functional enzymes is continuously increasing, necessitating the development of novel computational techniques [327]. Zou et al. [328] proposed mlDEEPre, a deep learning network specifically designed to predict the functionality of multi-functional enzymes. Using an auto label issuing threshold and a new transfer function linked with the interaction between several labels, mlDEEPre could predict multi-functional enzymes reliably and quickly. The proposed multi-label model surpassed all other approaches, as it correctly predicted 97.6% of all the observed primary categories in the test dataset with an SD of 0.27. The performance of SVM-NN with an accuracy of 84.7% seemed somewhat better than that of mlDEEPre (82.6%) and GA (80.8%), despite the fact that SVM-NN was worthy of predicting such infrequent class labels generated by unbalanced training samples. Comprehensive tests further demonstrated that mlDEEPre outperformed the other techniques in determining the kind of functional enzyme utilized, as well as the primary class prediction throughout many parameters. As mlDEEPre and DEEPre are flexible, mlDEEPre could be effortlessly merged into DEEPre, allowing the enhanced DEEPre to handle different functional predictions without human interference.

### 3.4.3 Drug discovery and toxicology

Emerging contaminants (ECs), which pose a serious risk to human health due to their detrimental effect on the endocrine system, include over eighty thousand endocrine-disrupting chemicals (EDCs). To determine the possible impacts of EDCs, numerous in vitro techniques have been developed, such as signaling pathway studies, ligand-binding protein, reporter gene experiments, and cell proliferation [329]. While in vitro methods are often more economical and quicker than in vivo trials, it is still impractical to analyze thousands of molecules in a reasonable amount of time [330]. This has resulted in the usage of alternative computational methods experiencing exponential growth. To anticipate the toxicological effects of chemicals, the QSAR model is a possible substitute for in vitro techniques.

A DL-QSAR model was evaluated by Heo et al. [331] to give predictions regarding the impacts of EDCs on the endocrine system, particularly estrogen receptor (ER) and sex-hormone binding globulin (SHBG). For the classification and forecasting of probable EDCs, DL-QSAR models were created, and three distinct DL algorithms, namely SAE, DBN, and DNN, were employed. The suggested models' performance was assessed using validation metrics by comparing the classification prediction models with traditional machine learning classifiers, such as LR,

SVM, and MLR. Results showed that the DL-QSAR algorithm outperformed traditional machine learning based QSAR models. Accordingly, the DNN-QSAR model adequately described most EDCs' qualitative responses to tests. Compared to the LR and SAE-QSAR models (accuracies of 86.49% and 89.91%, respectively), the DNN-QSAR model achieved an accuracy of 90%. As a result, DNN was more effective for evaluating qualitative responses since it could translate dense chemical identifiers into multidimensional space regions. Additionally, by overcoming multicollinearity and overfitting issues, DNN-QSAR exhibited great performance in computational chemistry. As a result, it was determined that DL can effectively utilize the qualitative characteristics of the EDCs.

Effective methods for assessing potential EDCs are majorly required. An evaluation platform was established by Zhang et al. [330] established an evaluation platform by employing three different machine learning models, such as SVM, linear discriminant analysis (LDA), and classification and regression trees (CART), to identify EDCs through estrogen receptors. The model was modified using 440 compounds, and 109 new compounds were added for the screening of EDCs. Their predictive capabilities were evaluated by contrasting the screening results with those anticipated by the models derived from classification. The most accurate model was found to use an SVM classifier, which correctly identified agonists at an accuracy level of 76.6% and antagonists with an accuracy of 75% on the test set, including an average predicted accuracy of 75.2%. The overall projected accuracy confirmed by the screening of the EDC assay was 87.57%, illustrating the effectiveness of a synthetic alert for EDCs with ER agonistic or antagonistic actions.

Drugs are often the cause of detrimental results, such as accidents, injuries, and enormous medical expenses. Clinicians can build suitable treatment plans and make effective judgments with the aid of accurate drug-drug interaction (DDI) predictions. Recently, numerous AI-based methods for DDI prediction have been proposed. However, the majority of currently used techniques give less consideration to possible connections between DDI processes and other multidimensional data, such as targets and enzymes. A multimodal DNN for DDI events prediction was suggested by Lyu et al. [332] in order to generate drug multimodal representations in a multimodal deep neural network (MDNN). To examine the complementary aspects of the drug's heterogeneous representations, a multimodal neuronal layer was also developed. Several multi-class classification evaluation metrics, such as F1 score, accuracy, and recall, were used to assess the prediction performance. The proposed model, MDNN, achieved the most stable performance and outperformed DDIMDL by 0.08% on accuracy, 7.1% on F1 score, 1.5% on precision, and 10% on recall. The MDNN model achieved an accuracy of around 99%, according to the comparative study with other models used in different journals, such as DDIMDL [333]. The MDNN model's superior performance can be ascribed to its exploration of both the cross-modality embedding representations of the heterogeneous data and the drug topological integrating expressions in the drug network. This effectively demonstrates how structural information and multimodal features can increase the prediction accuracy of drug-drug interaction irrespective of current drugs or newly-launched FDA-approved drugs and provides a solid, trustworthy basis for research on DDI prediction.

For the prediction of drug-induced liver injury (DILI), Yang et al. [334] constructed 48 classification models using a combination of six distinct molecular fingerprints. These models were developed using a DNN and seven different ML algorithms. In comparing the outcomes of the DNN and ML models, the top-performing model was determined to be the one that utilized DNN with ECFP_6 as its input. The model achieved a balanced accuracy of 0.680, an F1-score of 0.753, and an area within the receiver operating characteristic curve of approximately 0.713. The DNN models built using molecular fingerprints could serve as reliable and effective tools for assessing the risk of DILI in the early stages of developing novel medications.

Drug co-prescription can be safer and more productive if the effects of drug-drug interactions (DDIs) are accurately predicted [335]. There is still potential for improvement in prediction performance despite the many computational methods presented to anticipate the impact of DDIs [336]. These methods attempt to make it easier to find these interactions in vivo or in vitro. A deep learning model was proposed by Lee et al. [337] to more precisely predict the impact of DDIs, as well as to employ autoencoders and a feed-forward deep network to predict the therapeutic effects of DDIs. The experiments utilizing just GSP or TSP or combined GSP and TSP did not produce tests with satisfactory classification accuracy. However, integrating TSP and GSP improved classification accuracy to 97-97.5%. Additionally, the suggested model outperformed standard techniques like SVM (80-83%) and Random Forest (RF) (75-91%). The results showed that TSP and GSP improved prediction accuracy in comparison to SSP alone, and the autoencoder outperformed PCA in reducing the parameters of each profile. Hence, the proposed deep learning model provided a more precise prediction of DDIs and their therapeutic effects.

A summary of the reviewed studies on the uses of deep learning in various sectors is provided in Table 4. Based on our prior discussion, Table 5 represents the relationship between several fields in terms of DL applications and their benefits, specifically how DL techniques on vision, audio, and NLP help other concrete applications such as transportation, agriculture, bioinformatics, and ecology.

Table 4. Overview of the surveyed studies conducted on the applications of deep learning

| Applications | Algorithms/ Models | Objective | Outcome | Remarks | Ref. |
|---|---|---|---|---|---|
| Transportation Prediction | Attention based ConvLSTM, Bi-LSTM | Extracts fundamental properties of traffic flow data using hybrid and multiple layer architectures | The combination of attention ConvLSTM and Bi-LSTM performed better than the existing models. | They considered a relatively small and simple road network. Therefore, CNN and Bi-LSTM methods might be unable to completely utilize traffic flow's complex and dynamic properties. | [301] |
| | Attention-based CNN-LSTM | Traffic flow forecasting | The accuracy of the model was found 99% | Weather and other factors such as accidents and road closures can be factored into the model to enhance it. | [302] |
| | Deep stacked autoencoder to present features in a lower dimension | Predict the travel times | Showed better performance than applying Deep NN in the initial training data. | Wasn't capable of capturing traffic dynamics during heavy snow. | [303] |
| Agriculture | Transfer learning to train Deep Convolution NN | Plant leaf stress detection | Achieved 93% accuracy | For certain diseases (CBSD, BLS, GMD), using the leaflet rather than the full leaf increased diagnostic accuracy. However, using whole leaf photos enhanced accuracies for others (CMD and RMD). | [254] |
| | Multi-layer CNN | Distinguish between healthy and stressed mango leaves | Achieved 97.13% accuracy | Employing a new activation function in place of Softmax can improve CNN's performance | [338] |
| | CNN | Plant disease identification | Mildly diseased images were difficult to identify but accuracies were higher for other cases. | In all situations, a few hundred photos appeared to be sufficient to produce credible findings, but this quantity must be approached with caution. | [256] |
| | UNet-CNN | Classify and identify cucumber powdery mildew-affected leaves | CNN model segmented the sick powdery mildew on cucumber leaf pictures with a mean pixel accuracy value of 96.08%. | Lack of appropriate amount and diversity of the datasets. | [257] |
| | CNN | Fish species classification | Achieved accuracies of over 90% | General deep structures in the experiment should be fine-tuned to increase the efficacy to identify vital information in the feature space of interest, in order to reduce the requirement for vast volumes of annotated data. | [259] |
| Natural language processing | Collaborative adversarial network | Paraphrase identification | The model outperforms the baseline MaLSTM model | Shows excellent potential using the CAN for paraphrase identification | [175] |
| | RNN (LSTM) and several other models | Paraphrase identification | RNN (LSTM) outperforms the others | RNN shows significant performance in NLP | [176] |
| | Weighted Transformer | Machine Translation | Outperformed state-of-the-art methods | The model ignores the modeling of relations among different modules. | [339] |
| | Single Neural Machine Translation | Machine Translation | The method works reliably on Google scale production setting | The performance of zero-shot translation is frequently insufficient to be practical, as the basic pivoting strategy quickly outperforms it. | [177] |
| | Deep-attention model | Machine Translation | In comparison to the best methods currently available, deep attention performs exceptionally well | The technique could be implemented for other tasks like summarization and could adapt to more complex attention models. | [178] |
| | BiLSTM with CRF | Sentiment analysis (extracting aspect opinion target expression) | Outperformed the existing research | Unable to represent several aspects of sentences and did not investigate the explicit location contexts of words. | [182] |
| | LSTM | Sentiment Analysis | Outperformed the existing research | Unable to represent several aspects of sentences and did not investigate the explicit location contexts of words. | [183] |
| | Domain attention model | Multi-domain sentiment categorization | Evaluated on multiple datasets and showed better performance than the state-of-the-art techniques | Can pull out the most distinctive features from the hidden layers, reducing the number of labeled samples required. | [340] |

| | | | | |
|---|---|---|---|---|
| Weakly-supervised multimodal deep learning (WS-MDL) | Prediction of multimodal attitudes in tweets | Better performance than supervised and other weakly supervised models. | The order of the emotion icon levels might be further investigated, which may be included as a constraint to the proposed WS-MDL technique. | [341] |
| CNN, LSTM | Sentiment classification | Achieved accuracy of 87% | In future work, bag-of-word and word embedding techniques can be combined | [342] |
| ConvLstm (merged CNN and LSTM) | Sentiment analysis | Performed better than existing works with less number of parameters | Local information loss may be reduced, and long-term dependencies can be captured using the suggested design | [343] |
| Attention-based LSTM | Question answering | Better performance than baseline approaches | Modeling more than one aspect simultaneously with the attention mechanism would be an interesting addition to the experiment. | [187] |
| Hierarchical TF-IDF document retriever and a BERT document reader | Question answering | Significant improvement of the baseline techniques for the Arabic language | They introduced a dataset (ARCD). However, ARCD's questions were created with certain paragraphs in mind; without that context, they might seem ambiguous. | [188] |
| CNN, LSTM | Visual question answering | Will be helpful to study the reduction of the network model size. | Determining the rank is an NP hard problem in the low-rank decomposition, and their method is still constrained in this area by inserting hyper-parameters | [189] |
| Tree-LSTM | Visual question answering | Better performance than existing Hie and Deeper LSTM | The representative ability of the network could be improved in the future. | [190] |
| CNN, LSTM | Visual question answering | Outperformed the existing studies by 5% | The approach might be used on unstructured information sources, such as online text corpora, in addition to structured knowledge bases. | [344] |
| Reinforcement Learning technique and an encoder-extractor | Summarization | Performed better than baseline Bi-LSTM | The model may suffer from significant variance since they use an approximation in the Reinforcement Learning method training objective function | [194] |

| | | | | | |
|---|---|---|---|---|---|
| | | network architecture's RNN sequence model | | | |
| Biomedicine | Cascaded-CNN (C-CNN) | Prediction of alpha carbon atoms throughout the core structure of proteins | C-CNN outperformed (88.9%) the Phoenix-based structure construction method (66.8%) | Adding protein structural details for training the networks can enhance the model | [310] |
| | deepMiRGene (RNN+LSTM) | Prediction of structural characteristics of precursor miRNAs | deepMiRGene performed better, having accuracy between 88%-91% | The most important contribution was the elimination of rigorous manual feature development. | [311] |
| | DNN | Prediction of gene expression interference | Outperformed linear regression in 99.97% of the key genes | Demonstrated higher accuracy than the linear regression model | [317] |
| | FCNNs, CRF, RNN | Cerebral malignancy identification | The integration of FCNNs and CRF-RNN acquired an accuracy of 88% | 2D CNNs lack the essential capabilities to fully utilize 3D information from MR | [320] |
| | CNN | Identifications of kernels for gliomas segmentation in the images derived from MRI | 88%, 83%, 77% accuracy acquired | CNN model demonstrated to be legitimate in the Cerebral Tumor Classification database | [322] |
| Bioinformatics | DNN, RNN | Identification of AMPs | Best performance, having an accuracy of 91.01% | Eliminated the dependency on domain experts for feature creation by employing a deep network model | [323] |
| | DeepCpG | Single-cell DNA methylation prediction | Obtained accuracy of 83% | The strength of the DeepCpG extracting predictive sequence from large (<1000 bp) DNA sequence. | [324] |
| | EPIVAN | Identification of enhancer-promoter interactions | Accuracy ranging from 96.5% to 98.5% in different datasets | The model can be utilized in transfer learning | [325] |
| | mlDEEPre | Prediction of Multiple Enzyme Functions | 97.6% accuracy with a standard deviation of 0.27 | mlDEEPre can be effortlessly merged into DEEPre to handle functional predictions. | [328] |

| | | | | |
|---|---|---|---|---|
| Disaster Management Systems | CDNN algorithm | Flood catastrophe identification | Higher accuracy (93.2%) level than the current approach of DNN and ANN | The model can be improved with IoT-based devices using cutting-edge algorithms at each stage of flood identification | [278] |
| | CNN-based architecture | Early fire detection | High accuracy of from 89% to 99% | Capable of early-stage fire detection with high accuracy and in addition to providing an automated response | [281] |
| | SVM, LR, CNN classifier | CNN-based deep learning algorithm to classify the trending catastrophic topics from social media | SVM (63%-72%), LR (44%-60%), CNN (81%) | CNN took longer to learn than SVM and LR as there were more parameters to consider, making it difficult to employ for web-based learning | [51] |
| | RNN, CNN, MCA-based model | Disaster information management | Minimal accuracy 73% | MCA based model can include more textual and metadata to optimize the final categorization results | [345] |
| Drug discovery and toxicology | DNN-QSAR model (SAE, DBN, DNN) | Prediction of impacts of EDCs on the endocrine system, particularly (SHBG) and (ER). | The accuracy of DBN-QSAR was 90%, | DNN was more effective for evaluating qualitative responses | [331] |
| | SVM, LDA, CART | Identifying EDCs through the ER | Overall projected accuracy confirmed by the screening of EDC assay was 87.57% | The estrogenic activity of 109 compounds was predicted using the best model, SVM. | [330] |
| | CNN | Reaction prediction and Retrosynthesis | The model achieved a 95% retrosynthesis accuracy and a 97% response prediction accuracy. | The model demonstrated the extreme capability of ranking the real reactions precisely. | [346] |
| | Multimodal deep neural network (MDNN) | To build a connection between DDI and other targets and enzymes. | DDIMDL outperformed by achieving around a 99% accuracy rate | Demonstrated the impact of structural information and multimodal features on prediction accuracy | [332] |

| | | | | |
|---|---|---|---|---|
| | Autoencoder (SSP, TSP, GSP) | Predict the therapeutic effects of drug-drug interaction | classification accuracy between 97%–97.5% | Therapeutic implications of the DDIs predicted should be verified | [337] |
| Partial Differential Equations | 5-layer deep neural network | Building multi-physics/multi-scale modeling | For noise-free training data, the error in predicting unknown parameters was 0.023% and 0.006%, respectively. | Capable of accurately identifying the unknown parameters despite variations and the significant time interval between the two training images. | [347] |
| | Deep residual network (ResNet) | Solving SBVPs with high-dimensional uncertainty | DNNs predicted the mean with less than 1.35% relative $L_2$ error | Trained DNNs were found to transition effectively to inputs from non-distribution data | [348] |
| | CNN-based encoder-decoder network | Solving SPDE and uncertainty assessment tasks | 300 epochs and 200 epochs for PCS and DDS, respectively, along with 8 to 12 images | Surrogate model PCS outperformed data-driven DDS and demonstrated the capability of incorporating prediction uncertainty | [349] |
| Financial fraud detection | LTSM, RNN, ANN | Evaluating the effectiveness of deep learning techniques in financial fraud detection | LSTM technique achieved the best performance | Did not specify the network size and sensitivity | [269] |
| | LSTM | Credit card fraud detection | Detected suspicious financial activities and alerted the appropriate authorities with 99.95% accuracy | It can study even complex data structures and adjust to changed fraud trends dynamically. | [265] |
| | $H_2O$ framework | Credit card fraud detection | Enabled multiple algorithms to aggregate as modules, and their outputs can be integrated to improve the final output accuracy. | Adding more algorithms with equivalent formats and datasets can elevate this model. | [271] |
| Computer Vision | CNN | Generic object detection | The proposed method increased the RCNN's mean average precision from 31 to 50.3. It also exceeds GoogLeNet, the winner of the ILSVRC2014, by 6.1% | The addition of a def-pooling layer provides the model with a richer set of options for handling deformations and incorporating deep architectures. | [207] |

| | | | | |
|---|---|---|---|---|
| Semi-Supervised | Overcoming the crucial drawbacks of the object detection model | Translated the input data into more compact and abstract representations, which enhanced the model's convergence, stability, and performance | Allows dynamic modification of the model to the digital photogrammetry conditions. | [208] |
| Global adversarial deep CNNs | Face recognition model | Protocol attained 94.05% state-of-the-art verification accuracy | Can generate identity-distilled features and also extract concealed identity-dispelled features but lacks generalization | [213] |
| CNN and LSTM | Human activity recognition | Outperformed previous models by up to 9% | The framework is useful with similar sensor modalities, and it can combine them for enhanced performance. | [222][271][271][271][270] |
| LSTM | Decreasing the parameters and computational cost of the preceding action and activity recognition model | The overall accuracy of this project was 95.78% on the WISDM dataset | It was tested on enough datasets to guarantee its generalizability | [223] |
| Autoencoder | Building an autoencoder system with minimal supervision via face identities | The model was able to generate identity-distilled features and also extract concealed identity-dispelled features. | Did not specify the exact dependence on range resolution | [274][274][274][274][273] |
| ResNet based network | Human pose recognition | The network could forecast joint heatmaps and relative displacements of all key points for all people in real-time. | Notwithstanding its simplicity, it offered state-of-the-art outcomes against the challenging benchmark. | [229] |

| Ecology | Mask R-CNN | To compare the deep learning algorithms for determining fish abundance with human counterparts | The deep learning algorithm outperformed marine specialists by 7.1% and citizen scientists by 13.4% | Evaluated abundance with stable results and is more portable across survey locations than humans | [247] |
|---|---|---|---|---|---|
| | CNN | Tree defoliation identification | Identified tree defoliation 0.9% less accurately than a group of human experts | Can produce errors during complex situations due to data deficiency | [248] |
| | DNN and Transfer learning | To automate characteristic recognition and extraction. | The top-one species identification accuracy was 80%, while the top-five accuracy was 90% | It is not possible anymore owing to vast differences in visual appearance | [237] |
| | CNN | Directly identifying ant genera from the profile, head, and dorsal views of ant photos. | Gained over 80% accuracy in top-1 classification and over 90% in top-3 classification while reducing total classification error | Contributes novel understanding of ensembles for multi-view structured data and transfer learning processes for probing commonalities in multi-view CNNs | [251][271][271][271][270] |
| Fluid Dynamics | Deep CNN | To scrutinize heat transport characteristics of turbulent Rayleigh–Bénard convection. | Simplified the complicated 3D superstructure in the midplane layer | Did not provide information on the network's efficiency at lower Rayleigh numbers | [285] |
| | Physics-guided neural networks (PGNN) | To forecast turbulent flow | Showed significant improvements in error prediction over state-of-the-art benchmarks | Precisely simulate the turbulent kinetic energy field and spectrum that are vital for accurate turbulent flow prediction. | [286] |
| | DLROM and POD | Describing the conservation of mass and momentum in fluids | Comparison with earlier ROMs clarified that the DLROM has better potentiality in the prediction | The consequences of applying this approach to transitioning parametric constraints, such as real-time response to natural disasters, are uncertain. | [47] |

| | | | |
|---|---|---|---|
| ROMs and POD | To conduct a comparative analysis using three ROMs applied to a biological model | A comparison of POD-DEIM and Gappy POD solutions revealed similar levels of accuracy | The model did not converge when the number of MPE points was low [287] |
| POD and ROM | Using an efficient adjoint technique to optimally collect targeted observations | When compared to the high fidelity model, the size of the problem is decreased by a factor of 200 | It ensures that the sensors are positioned at the optimum distance from one another. [350] |

Table 5. How deep learning techniques on vision, audio, and NLP help the other concrete applications

| Application area | DL techniques | Benefits and applications |
|---|---|---|
| Transportation | Object detection | - Enhancing autonomous driving by identifying pedestrians, vehicles, and road signs<br>- Real-time surveillance for safer navigation<br>- Traffic flow analysis and optimization |
| | Image segmentation | - Identifying road lanes and obstacles for self-driving cars<br>- Accurate mapping and localization using satellite imagery |
| Agriculture | Crop monitoring | - Identifying crop diseases, pests, and nutrient deficiencies through image analysis<br>- Precision agriculture and yield prediction |
| | Object recognition | - Differentiating between various plant species and weeds for targeted intervention |
| Disaster management | Image analysis | - Aerial imagery interpretation to assess disaster extent and damage<br>- Identifying trapped survivors in disaster areas |
| | Sentiment analysis | - Analyzing social media data for real-time disaster response and understanding public sentiment |
| Drug discovery | Molecular structure | - Predicting molecular interactions and drug-binding sites for drug design<br>- Accelerating the drug discovery process |
| | Image analysis | - Analyzing medical images to identify potential drug candidates and understand their effects |
| Toxicology | Toxicity prediction | - Predicting toxicity of chemicals and compounds, aiding risk assessment<br>- Reducing animal testing through computational models |
| Bioinformatics | Sequence analysis | - Analyzing genetic data for disease prediction and personalized medicine<br>- Identifying genetic markers for various conditions |
| | Protein structure | - Predicting protein structures to understand their functions and interactions<br>- Advancing drug discovery and understanding diseases |
| Ecology | Image classification | - Identifying and tracking wildlife species for biodiversity monitoring<br>- Detecting habitat changes and invasive species |
| | Sound classification | - Monitoring ecosystem health by analyzing animal calls and environmental sounds |

| Fluid dynamics | Image analysis | - Analyzing fluid flow patterns in images or simulations |
| | | - Identifying turbulence and vortices for better predictions |
| | Data analysis | - Processing large volumes of flow data for insights and predictions |
| Civil engineering | Image recognition and analysis | - Analyze aerial or satellite images to monitor infrastructure |
| | | - Detect changes or potential concerns in construction areas |
| | | - Evaluate the state of bridges and roads |
| | Noise monitoring | - Assessing the effects of construction activities on neighboring communities |
| | | - Assuring attention to noise regulations |
| | | - By examining acoustic signals, structural anomalies can be detected |

***Summary:*** Applications of deep learning demonstrate the adaptability and revolutionary power of deep learning in solving complicated problems and advancing technology in many fields. Deep learning has strengths as well as shortcomings when applied to different fields. Although sophisticated audio analysis and precise voice recognition are two of the domain's strongest points, dealing with various accents and ambient noises can be a real challenge. Deep learning excels at optimizing routes and predicting traffic, but its limitations may become apparent in highly urbanized areas or situations where traffic is constantly changing. Nevertheless, these strengths are useful for transportation prediction. For instance, traffic forecasting can be made with 99% accuracy using a CNN-LSTM model based on an attention mechanism. Precision farming has strengths in agriculture, such as crop monitoring and yield prediction. For example, the multi-layer CNN model can reveal 97.13% accuracy in classifying mango leaves affected by fungal disease. However, adapting models to diverse agricultural landscapes is still a challenge. When it comes to cultural differences and complex contexts, natural language processing applications may fall short, but they excel when it comes to sentiment analysis and machine translation. Genomic analysis and drug discovery are two areas where DL excels in biomedicine and bioinformatics. In 99.97% of targeted genes, deep learning can outperform linear regression in terms of error reduction. For prolonged enhancer-promoter interactions (EPIs) that can be predicted solely from genomic sequences, the EPIVAN deep learning model achieves an accuracy of 96.5% to 98.5%. However, there are a few obstacles, such as the difficulty in understanding complicated models and a lack of data in specific fields.

While alert systems are helpful for disaster management, they may fail to adequately anticipate truly unique occurrences. Toxicology and drug discovery use deep learning to save time and effort, but it's not always easy to predict complicated biological interactions. Though deep learning is an efficient solution to partial differential equations, it can be difficult to figure out the physical meaning of black-box models. Since adversarial attacks are possible, financial fraud detection depends on deep learning's anomaly detection strengths. With an accuracy of 99.5%, the deep learning model LSTM is capable of identifying fraud-related credit card activities and notifying the relevant authorities. Despite their usefulness, computer vision applications like object detection and image recognition can be influenced by biases and necessitate massive labeled datasets. Deep learning has many applications in ecology, but it can be difficult to generalize results to other ecosystem types. Similarly, fluid dynamics models benefit from deep learning, but they still have a ways to go before they can fully capture the complexities of real-world phenomena. Interpretability, generalizability, and domain adaptability are three areas where deep learning applications still face obstacles despite their many strengths.

## 4. Challenges and benefits of deep learning applications

Deep learning is an incredibly important computational tool primarily due to its accuracy in data prediction and analysis. Since DL does not require the use of previously processed data, the input of raw data can be computed through branched layers that separately analyze the data and represent it to the next layer for further processing [351]. Deep learning is useful when dealing with enormous amounts of data since it can extract information, eliminating the need for data training and the associated computational expenditures. It is also notable for its expression and optimization capabilities, which make it so effective at processing data without training [352]. As the volume and complexity of available data continue to rise exponentially, deep learning is gaining recognition as a crucial tool for more efficient data collection and analysis [353].

As demonstrated in the literature, deep learning applications offer an extensive range of advantages across diverse domains. Speech and audio processing techniques demonstrate exceptional proficiency in accomplishing tasks such as voice recognition, facilitating enhanced communication interfaces with superior accuracy. Deep learning models in transportation prediction offer accurate forecasts, optimizing traffic flow and improving overall efficiency. These applications enhance agriculture as they facilitate precision farming by monitoring crops and detecting diseases, resulting in higher crop yields. Deep learning is utilized in natural language processing applications to achieve sophisticated language comprehension, which consecutively enhances machine translation and sentiment analysis by providing more nuanced results. It expedites the examination of intricate genomic data in biomedicine and bioinformatics, facilitating drug discovery and disease prognosis.

Several challenges of deep learning applications are common across various fields. Unfortunately, deep learning models are not always practical in environments with limited computing resources because of how much power they consume. Because of its independence from training data, deep learning demands rigorous data collection for proper analysis and processing of such a large number of data. Hence, medical, research, healthcare, and environmental data are challenging for large-scale data compilation, reducing the efficacy of deep learning [354]. Data quality and structure are particularly concerning, as data from health, research, and environmental studies are highly heterogeneous, full of ambiguity and noise, and often incomplete, which frequently presents problems for the model. Another issue with deep learning is that it typically assumes inputs to be static vectors and cannot readily incorporate time variables as inputs. Due to the complex signal relationship across time, samples are irregular, which influences the model's performance, particularly when dealing with health data [355–360].

Domain complexity, such as diverse data and insufficient information, and black boxes, the complexity of algorithms that aren't understood, offer challenges for the system and inhibit the growth of data comprehension [351, 361, 362]. In important domains like biomedicine, where comprehending the decision-making approach is essential, the interpretability of these models continues to be an issue. However, the interpretability of complex models can be significantly improved in practice through the implementation of knowledge distillation and attention mechanisms. Knowledge distillation involves the transfer of insights from a complex model to a simpler model, achieving a balance between predictive performance and interpretability. Attention mechanisms, inspired by human cognitive processes, facilitate the focus of models on specific elements of input data, thereby providing practitioners with valuable insights into the process of decision-making [363]. In practical implementation, these solutions are seamlessly integrated into training and evaluation pipelines, bridging the gap between model complexity and human understanding. The real-world impact lies in fostering trust and confidence in AI systems, as interpretable models enhance user comprehension and enable practitioners to identify biases, understand model decisions, and troubleshoot effectively, contributing to the overarching goals of transparent and trustworthy artificial intelligence [364].

The intelligibility of complicated biological interactions might be a problem for drug development applications. When detecting financial fraud, deep learning models are vulnerable to adversarial attacks. Despite these obstacles, deep learning still drives innovation and solves complex problems in many fields. Some studies have found that using multimodal data improves the accuracy of the results, while others report that the heterogeneous nature of the data makes it difficult for the program to implement the necessary mechanisms [365–368]. It is also challenging for deep learning algorithms to overcome the problem of label omission in many datasets [56]. The primary challenges and advantages associated with the integration of deep learning techniques within the investigated fields are presented in Table 6.

Table 6. Key challenges and benefits of implementing deep learning in the surveyed disciplines

| DL application fields | Challenges | Benefits |
| --- | --- | --- |
| Natural Language Processing (NLP) | – Imperfect data can lead to skewed results. <br> – Computational resources are a restricted commodity when dealing with large models. <br> – Some applications may struggle due to a lack of explainability. | – Computes at or near the state-of-the-art on several NLP tasks. <br> – Allows for the use of chatbots, sentiment analysis, and the translation of languages. <br> – Supports massive amounts of textual information. |
| Computer Vision | – Annotating and labeling data can take a lot of time. <br> – The processing power required by some deep learning models may be high. | – The ability to recognize faces and other objects in pictures is unlocked. <br> – Reached the level of performance seen in humans in some situations. <br> – It has potential uses in robotics and self-driving cars. |
| Speech and Audio Processing | – Needs a lot of labeled data to work. <br> – Extremely computationally demanding, which can limit its applicability to devices capable of real-time processing. | – Increased accuracy in speech recognition. <br> – Useful for a wide range of audio applications, including speech synthesis and noise cancellation. <br> – Capable of handling massive and intricate data sets. |
| Disaster Management | – Depending on the quantity and quality of available data.<br>– The stability of the model under severe circumstances is a matter of concern.<br>– Issues about the morality of data collection and analysis. | – Helpful for early warning systems, estimating damages, and allocating available resources.<br>– Uses satellite and sensor data for tracking emergencies.<br>– Facilitates quicker responses and better decisions. |
| Agriculture | – Needs knowledge and data unique to the given domain.<br>– Models may have limited applicability outside a specific geographical area or crop type. | – Facilitates disease diagnosis and crop yield forecasting.<br>– Drone footage and sensor information improve precision farming.<br>– Be able to make the best use of available resources. |
| Transportation Prediction | – The use of location data raises issues of data privacy.<br>– Deep learning algorithms may be difficult to interpret.<br>– Calls for a lot of processing power. | – Capable of simulating and predicting complex traffic patterns.<br>– Makes use of current information to improve forecasts.<br>– Facilitates the use of various transportation modes. |
| Biomedicine and Bioinformatics | – Handling sensitive medical information raises ethical and privacy concerns.<br>– Model interpretability is especially important for life-saving medical applications. | – Facilitates the study of drugs, proteins, and genomes.<br>– Reduces the time it takes to analyze medical images and diagnose diseases.<br>– Capable of dealing with complex biological data sets. |
| Drug Discovery and Toxicology | – Drug discovery with limited interpretability carries some inherent risks.<br>– Adopting AI-driven drug discovery faces regulatory obstacles. | – Finds potential drug candidates faster and helps predict their safety.<br>– Supports substantial chemical databases.<br>– Reduces the time and costs needed for experiments. |
| Fluid Dynamics | – The computational demands of 3D simulations are very high.<br>– Black-box models have limited knowledge of the underlying physical processes.<br>– Issues with precision and consistency can arise in a variety of contexts. | – Facilitates modeling and prediction of complicated fluid dynamics.<br>– Increases the speed at which CFD simulations can be run.<br>– Helpful in improving engineering and aerodynamic designs. |
| Partial Differential Equations | – Computer resources for 3D simulations are particularly demanding.<br>– The stability and precision of deep learning PDE solvers are not always easy to achieve. | – Specializes in fluid dynamics and solving complex PDEs numerically.<br>– Allows for varied simulations and models.<br>– Ability to enhance engineering designs. |

| Ecology | – Remote areas present unique challenges for data collection. | – Assists in the tracking and preservation of biodiversity. |
| | – The ability to transfer models between ecosystems varies. | – Forecasts the future results of current environmental changes. |
| | | – Uses sensor and satellite data for ecological study. |
| Financial Fraud Detection | – Training on large historical datasets is necessary. | – Explores financial data for intricate patterns. |
| | – There is the potential for model drift and false positives/negatives. | – Facilitates better transaction tracking in real time. |
| | | – Enables more precise fraud detection. |

## 5. Identified research gaps and future directions

### 5.1 Research gaps

Although deep learning has achieved notable progress across numerous fields, there remain several domains that continue to be investigated and developed further or have not yet reached their complete understanding:

i) For effective training, deep learning models frequently require vast quantities of labeled data. In domains where it is challenging to obtain large labeled datasets, such as specific medical disciplines or niche industries, the investigation of deep learning adaptations for regimes with limited data remains an unresolved challenge.

ii) Deep learning models, specifically deep neural networks, are frequently referred to as "black boxes" on account of their complicated architectures. Ongoing efforts are directed towards improving the interpretability and explainability of these models, with the ultimate goal of fostering confidence in their decisions, particularly in important domains such as finance and healthcare.

iii) Transfer learning has demonstrated efficacy in general tasks; however, further investigation is needed regarding the adaptation of pre-trained models to precisely match the characteristics and demands of specialized domains. Customizing pre-trained models to fit particular domains may result in improved performance and broader applicability.

iv) Deep learning models are vulnerable to adversarial attacks, in which even minor perturbations in the input can result in inaccurate predictions. Ongoing attention is being given to enhancing the resilience and security of DL models, particularly in domains where safety is paramount, including healthcare and autonomous vehicles.

v) The efficient management of temporal dynamics and sequential data presents an immense challenge. Time-series data that effectively captures long-term dependencies and patterns requires enhanced architectures and methodologies.

vi) An expanding field of study addresses ethical issues and prevents biases in deep learning models. Ongoing investigation is necessary to guarantee fairness, accountability, and transparency in model predictions.

vii) The seamless and successful integration of information from multiple modalities, including text, images, and audio, is an area of ongoing development known as multimodal integration. Ongoing research struggles to develop models capable of utilizing a variety of data types in order to achieve a complete understanding.

viii) The investigation of how to implement deep learning models on devices with limited resources, such as edge devices or Internet of Things devices, while simultaneously ensuring optimal performance and energy efficiency, is a current and dynamic field of study.

## 5.2 Potential avenues of future research

The potential of deep learning applications in the future is immense, presenting numerous promising avenues for scientific research and advancement. A summary of future directions for the application of deep learning across the various disciplines investigated in the present review is presented in Table 7.

Table 7. Prospects for the future application of deep learning across various fields of study

| Disciplines | Future directions |
|---|---|
| Speech and audio processing | – Advances in deep learning about resilient speech and audio processing might entail the construction of models capable of more efficiently managing a range of accents, dialects, and environmental noises.<br>– Investigating unsupervised learning techniques for learning representations in audio signals offers an additional prospective field of study. |
| Transportation prediction | – Construct models that aim to optimize traffic management, dynamic traffic flow adaptation, and real-time data integration.<br>– Investigate the potential for integrating deep learning with emerging technologies, including autonomous and connected vehicles. |
| Agriculture | – Address issues caused by a lack of data in specific agricultural settings and continue to build models for precision farming decision support, crop disease identification, and yield prediction.<br>– Consider incorporating IoT data and remote sensing for a more thorough examination. |
| Natural language processing | – Enhance understanding of cultural variations, idiomatic expressions, and context by models.<br>– Investigate developments in contextual embeddings, improve reasoning capabilities, and construct frameworks to facilitate understanding across different languages and cultures. |
| Biomedicine and Bioinformatics | – Develop models for the integration of multi-omics data, personalized medicine, and the discovery of new drugs.<br>– In addition to concentrating on interpretable models for clinical decision support, investigate the application of deep learning to understand complicated biological interactions. |
| Disaster management | – Construct decision-making models in real time that effectively incorporate a wide range of data sources, such as social media and satellite imagery.<br>– Improve upon existing models to facilitate comprehensive disaster risk assessment, adaptive resource allocation, and early alert systems. |

| Drug discovery and Toxicology | – Explore the potential of deep learning applications to enhance the precision of drug toxicity predictions, optimize drug design processes, and fully understand complicated molecular interactions. |
| | – Examine the potential of generative models in the generation of novel pharmaceutical compounds. |
| Partial differential equations | – Enhance the stability of deep learning-based solvers, develop more efficient architectures for numerical simulations, and expand the applicability of these methods to high-dimensional problems. |
| | – Analyze the application of neural networks based on physics to the solution of complex equations. |
| Financial fraud detection | – Improve the interpretability of the model and its ability to adapt to evolving fraud patterns. |
| | – Investigate how the integration of explainable AI, anomaly detection, and reinforcement learning can enhance the robustness of financial transaction fraud detection. |
| Computer vision | – Enhance adaptability against adversarial attacks, concentrate on achieving precise image recognition, progress models for comprehending visual relationships, and investigate unsupervised learning techniques. |
| | – Deep learning's application to 3D vision, scene comprehension, and object tracking should be investigated. |
| Ecology | – Tackle the obstacles associated with the scarcity of labeled data in ecological studies. |
| | – Construct models comprising capabilities for identifying species, monitoring ecosystems, and assessing biodiversity. |
| | – Consider the integration of environmental monitoring technologies with deep learning. |
| Fluid dynamics | – Develop neural architectures that capture complicated fluid behaviors, optimize for efficient simulation and turbulence modeling, and scale up applications to real-world scenarios. |
| | – Explore the application of deep learning in aerodynamics, fluid-structure interaction, and optimization of fluid flow problems. |
| Civil Engineering | – Develop predictive models that scrutinize sensor data derived from structures such as roads, bridges, and deterioration. |
| | – Developing more sophisticated computer vision systems and algorithms for reinforcement learning for use by autonomous construction equipment may be the subject of future studies. |

## 6. Conclusions

To effectively forecast and analyze data at all levels, deep learning uses cutting-edge methods, combining ideas from machine learning, deep neural networks, and artificial intelligence to construct models that utilize data representations at each level. This review investigated numerous deep learning frameworks, deep learning application pathways across diverse fields, and the obstacles and technical issues associated with using deep learning. Research gaps and potential future directions for deep learning applications were also addressed. Research domains might specialize in addressing specific difficulties and issues by analyzing the common obstacles encountered by deep learning applications across numerous fields. Data volume, quality, modeling, domain complexity, and representation are some issues that deep learning applications have in common with other disciplines. In order to facilitate the efficient experimentation and implementation of deep learning algorithms, deep learning frameworks provide developers and researchers with essential libraries and tools. This, consequently, promotes innovation and breakthroughs across various disciplines. TensorFlow and PyTorch, for example, are widely recognized for their flexibility and support from the community. These frameworks exhibit exceptional performance in the domains of computer vision, bioinformatics, and natural language processing. In the fields of financial fraud detection and drug discovery, TensorFlow, PyTorch, and Keras are dominant due to their robust libraries.

Demonstrating its revolutionary potential and adaptability, deep learning advances technology in many fields while solving complex problems. For instance, in the fields of biomedicine and bioinformatics, deep learning is particularly proficient in the areas of drug discovery and genomic analysis. In terms of error reduction, deep learning can outperform linear regression in 99.97% of the targeted genes. The EPIVAN deep learning model can achieve an accuracy of 96.5% to 98.5% for prolonged enhancer-promoter interactions (EPIs) that can be predicted solely from genomic sequences. However, deep learning models use much power, so they aren't always practical in places with limited computing power. In addition, domain complexity, characterized by a wide range of data and inadequate information, along with black boxes, which refer to algorithms that lack understanding, pose challenges for the system and impede the development of data comprehension.

Deep learning applications can be developed by studying sophisticated architectures, focusing on ethical and interpretability issues, incorporating developing technologies like quantum computing, and expanding deep learning applications. When processing massive amounts of data, it is best to employ a gated architecture, such as GRU units, for extracting persistent information. In a neural network designed for multimodal learning, some neurons are used for all tasks, while others are trained to perform specific tasks. The irregularity problem in temporal sequence similarity measurement is addressed by a suggested approach that uses dynamic time warming. For labeling challenges, data can be implicitly labeled by applying acquired knowledge to fresh datasets for the same task or by utilizing an autoencoder for a variant architecture such that transfer learning can be accomplished. Methods such as knowledge distillation, which condenses the data learned from a complex model to make it simpler and easier to execute, and attention mechanisms, which use an understanding of historical data to predict results, are also being used to enhance interpretability. Recent advances in quick solutions to current difficulties point to a promising future for deep learning techniques in multi-field applications. Future research should focus on better understanding the issues related to the construction of adequate datasets for deep learning models, including but not limited to data quality, volume, domain complexity, and privacy.

**Competing Interests**

The authors declare no conflict of interest.

## Authors contribution statement

SFA: Conceptualization, paper writing, and paper revision; MSBA: Paper writing, investigation; MK: Paper writing, formal analysis; SA: Paper writing, investigation; SJR: Paper writing, formal analysis; AA: Paper writing, resources; AHG: Review and editing; supervision.

## Ethical and informed consent for data used

Since no data were used in the study, there is no ethical or informed consent.

## Data Availability and access

No data has been used or produced to support the current study.